\documentclass[journal]{IEEEtran}
\usepackage{url}
\usepackage{cite}
\usepackage{graphicx} 
\usepackage{amsmath}
\usepackage{amssymb}
\usepackage{booktabs}
\usepackage{xcolor}
\usepackage{float}
\usepackage{afterpage}
\usepackage{array}
\usepackage[compatibility=false]{caption}
\usepackage{subcaption}
\usepackage{amsmath}
\usepackage{algorithmic}
\usepackage{array}
\usepackage{amsmath}
\ifCLASSOPTIONcompsoc
 \usepackage[caption=false,font=normalsize,labelfont=sf,textfont=sf]{subfig}
\else
 \usepackage[caption=false,font=footnotesize]{subfig}
\fi

\usepackage{stfloats}

\ifCLASSOPTIONcaptionsoff
 \usepackage[nomarkers]{endfloat}
\let\MYoriglatexcaption\caption
\renewcommand{\caption}[2][\relax]{\MYoriglatexcaption[#2]{#2}}
\fi

\usepackage{multirow} 

\newcommand{\etal}{\textit{et~al.}}

\def\BibTeX{{\rm B\kern-.05em{\sc i\kern-.025em b}\kern-.08em
    T\kern-.1667em\lower.7ex\hbox{E}\kern-.125emX}}
\begin{document}

\title{Out-of-Sight Embodied Agents: Multimodal Tracking, Sensor Fusion, and Trajectory Forecasting}


\author{Haichao~Zhang,~\IEEEmembership{Student Member,~IEEE,} 
        Yi~Xu,~\IEEEmembership{Student Member,~IEEE,}
        and Yun~Fu,~\IEEEmembership{Fellow,~IEEE}%
        \thanks{Haichao Zhang and Yi Xu are Ph.D. candidates in the Department of Electrical and Computer Engineering, Northeastern University, Boston, MA 02115, USA. Email: \texttt{zhang.haich@northeastern.edu}, \texttt{xu.yi@northeastern.edu}}%
        \thanks{Yun Fu is a Professor in the Department of Electrical and Computer Engineering and the Khoury College of Computer Sciences, Northeastern University, Boston, MA 02115, USA. Email: \texttt{yunfu@ece.neu.edu}}}

\markboth{IEEE Transactions on Pattern Analysis and Machine Intelligence}%
{Shell \MakeLowercase{Zhang \textit{et al.}}: Out-of-Sight Embodied Agents}

\IEEEpubid{10.1109/TPAMI.2026.3676710~\copyright~2026 IEEE}

\maketitle
\IEEEpubidadjcol

\begin{abstract}

Trajectory prediction is a fundamental problem in computer vision, vision-language-action models, world models, and autonomous systems, with broad impact on applications including autonomous driving, robotics, and surveillance. Most existing approaches assume observations are complete and relatively clean, and thus do not adequately address out-of-sight agents or the intrinsic noise in sensing modalities (e.g., sensor measurements) caused by restricted camera coverage, occlusions, and the lack of ground-truth denoised trajectories. These factors introduce substantial safety concerns and reduce the robustness of trajectory prediction in practical deployments.
In this extended study, we introduce major improvements to Out-of-Sight Trajectory (OST), a new task aimed at predicting noise-free visual trajectories of out-of-sight objects from noisy sensor observations. Based on our prior work, we expand the setting of Out-of-Sight Trajectory Prediction (OOSTraj) from pedestrians to both pedestrians and vehicles, thereby increasing its relevance to autonomous driving, robotics, and surveillance scenarios. Our improved Vision-Positioning Denoising Module utilizes camera calibration to construct a vision-position correspondence, mitigating the absence of direct visual cues while enabling effective unsupervised denoising of noisy sensor signals.
Extensive experiments on the Vi-Fi and JRDB datasets demonstrate that our method achieves state-of-the-art results for both trajectory denoising and trajectory prediction, with clear gains over prior baselines. We further provide comparisons against classical denoising techniques, including Kalman filtering, and adapt recent trajectory prediction models to this setting, establishing a stronger and more comprehensive benchmark. To the best of our knowledge, this is the first work to incorporate vision-positioning projection to denoise noisy sensor trajectories of out-of-sight agents, opening new directions for future research in this area. The code and preprocessed datasets are available at 
\url{https://github.com/Hai-chao-Zhang/OST}.

\end{abstract}


\section{Introduction}
\label{sec:intro}

\begin{figure}[t]
\IEEEpubidadjcol
\centering
\begin{minipage}[b]{0.49\linewidth}
    \centering
    \includegraphics[width=1.0\linewidth]{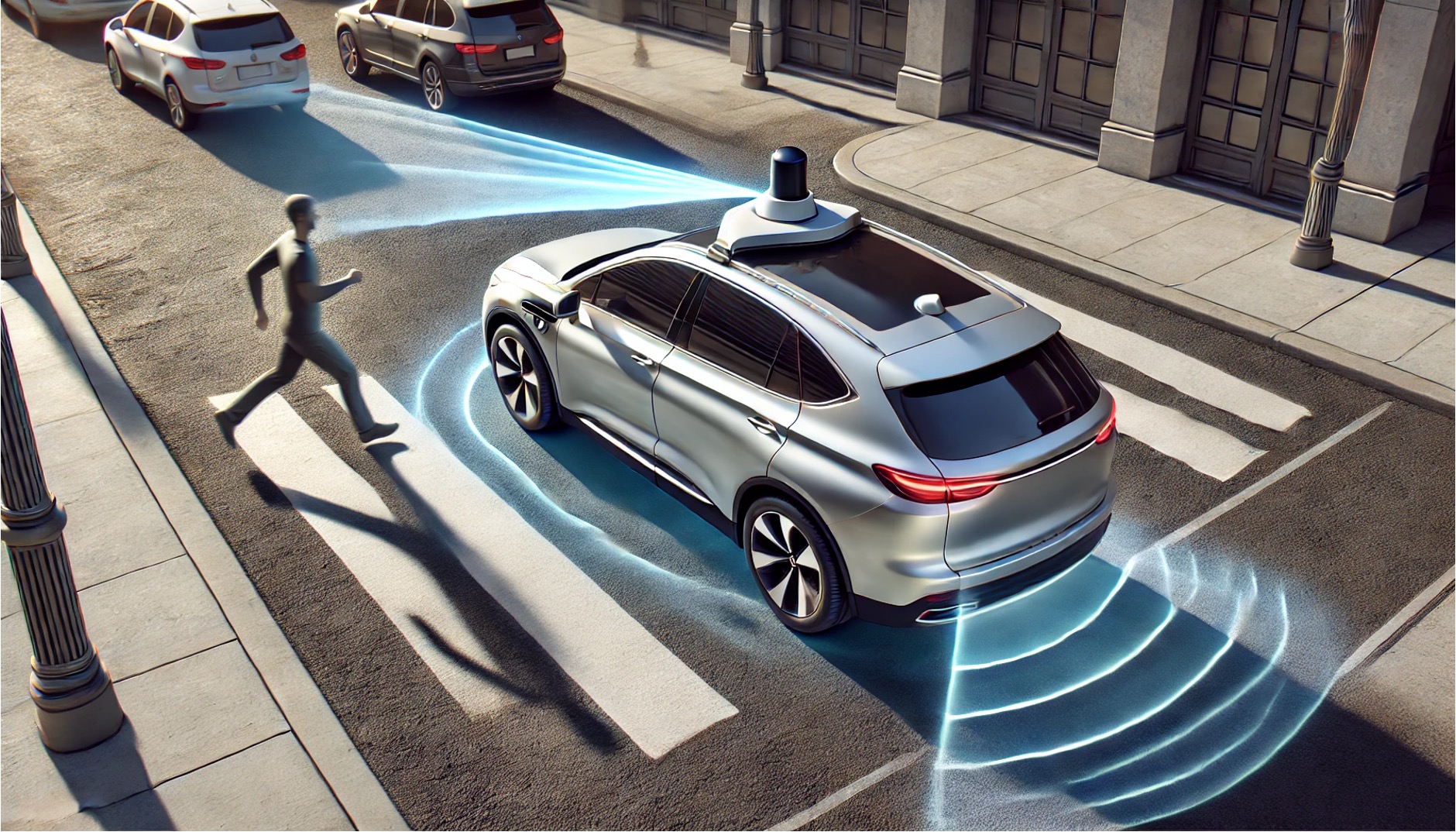}
    \caption*{(a)}
\end{minipage}
\hfill
\begin{minipage}[b]{0.49\linewidth}
    \centering
    \includegraphics[width=1.0\linewidth]{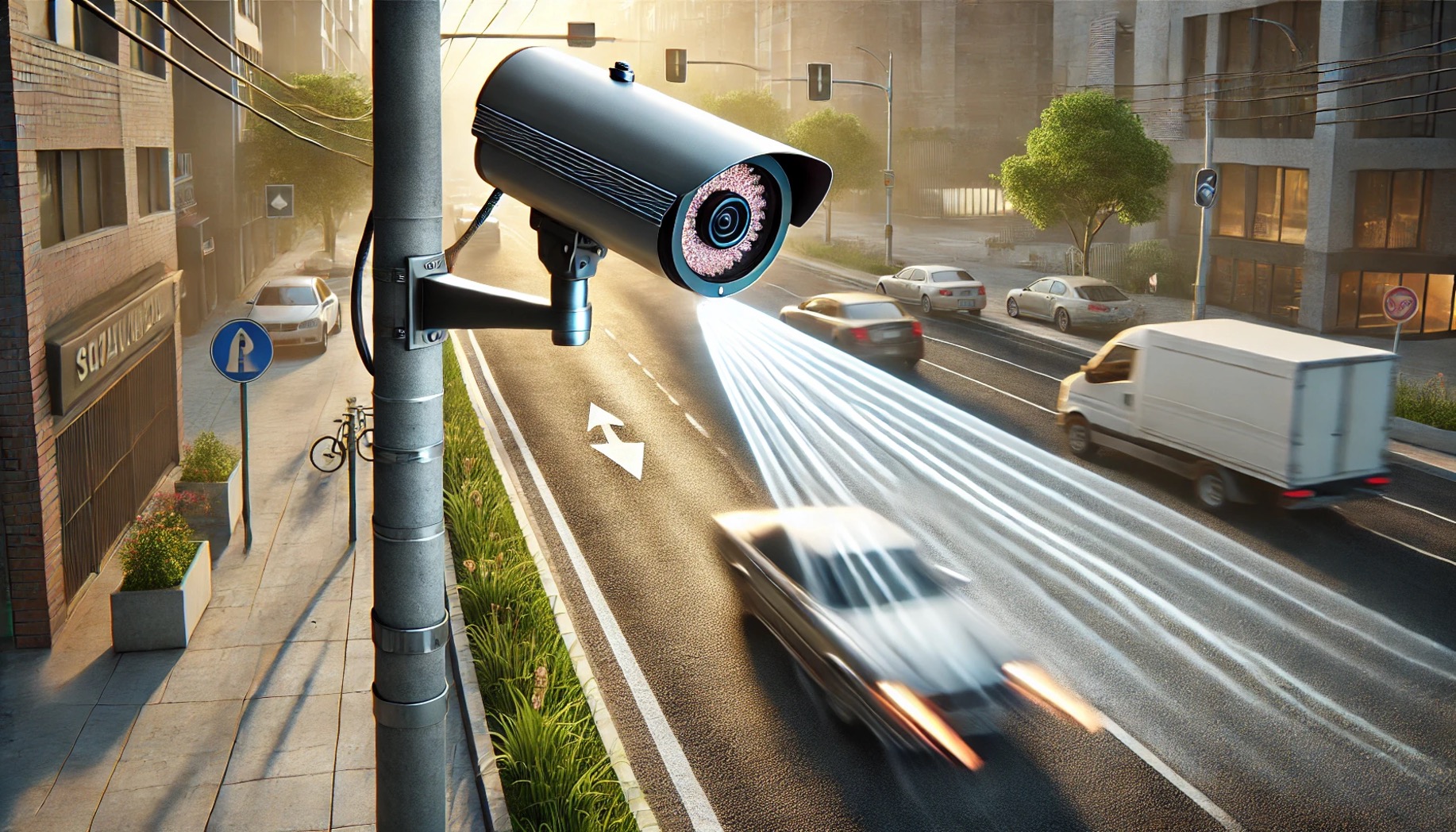}
    \caption*{(b)}
\end{minipage}
\vspace{10pt}
\begin{minipage}[b]{0.49\linewidth}
    \centering
    \includegraphics[width=1.0\linewidth]{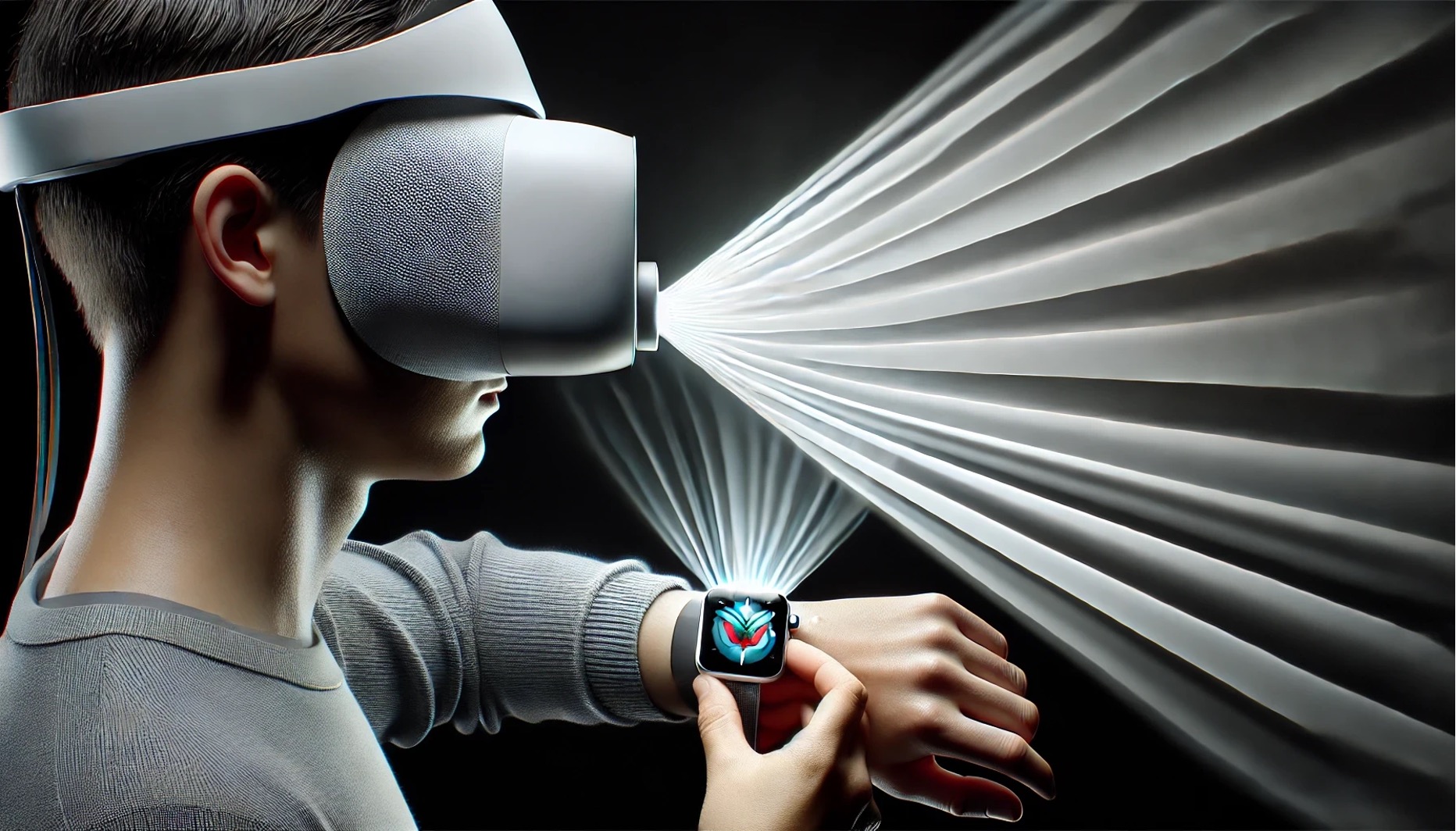}
    \caption*{(c)}
\end{minipage}
\hfill
\begin{minipage}[b]{0.49\linewidth}
    \centering
    \includegraphics[width=1.0\linewidth]{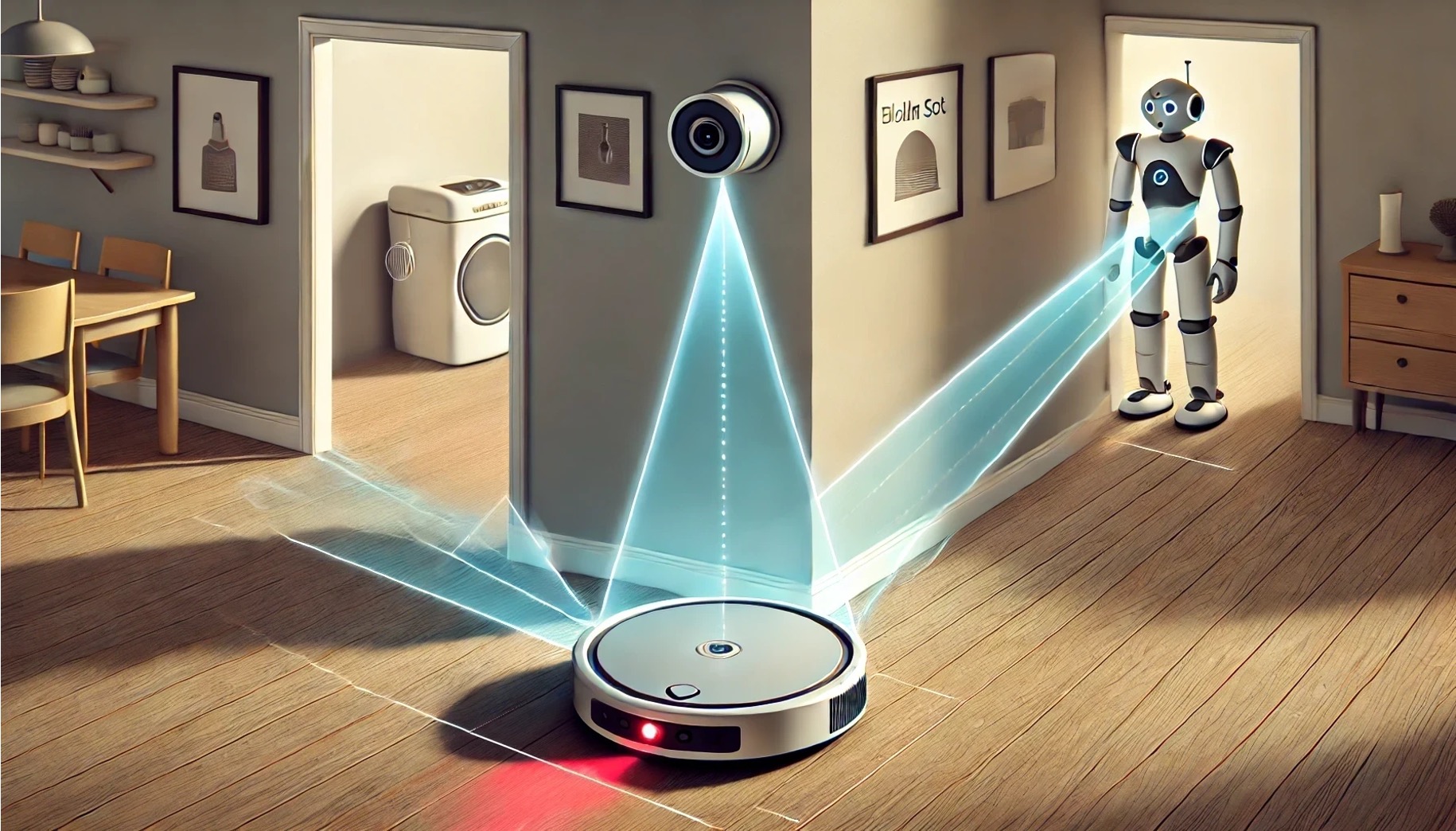}
    \caption*{(d)}
\end{minipage}
\caption{
Illustrations of typical scenarios where out-of-sight trajectory prediction is critical. 
(The lights in the figure indicate the field of view of each embedded camera.) 
(a) Autonomous vehicles face challenges with blind spots caused by limited sensors and cameras, often missing out-of-sight pedestrians or vehicles. Out-of-sight trajectory prediction aids in detecting and forecasting their future locations.
(b) Surveillance cameras with fixed angles frequently fail to capture vehicles entering or leaving their field of view. Out-of-sight trajectory prediction enables tracking of vehicles outside these restricted views.
(c) Head-mounted and ego-centric vision systems (e.g., AR/VR-style hand tracking) can lose track of hands or controllers when they move outside the camera field of view. While we do not evaluate on VR-specific datasets in this paper, out-of-sight trajectory prediction offers a general mechanism that could help maintain continuity under such out-of-view events.
(d) Robotics systems encounter occlusions where trajectories of IoT devices or other robots move out of sight. Out-of-sight trajectory prediction combines data from various viewpoints to track the positions and future paths of robots and IoT devices. 
\emph{Note:} Subfigure (c) is provided as a motivating example. Our experimental evaluation is conducted on JRDB and ViFi, and we do not include VR-specific datasets.
Figures generated with~\cite{ramesh2021dalle}.
}
\label{fig:scenarios}
\end{figure}

\IEEEPARstart{T}{rajectory} prediction is a fundamental capability for computer vision, vision-language-action models, world models, and autonomous systems, supporting applications such as autonomous driving, surveillance, and robotics (Fig.~\ref{fig:scenarios}). 
In real-world deployments, however, predictions must often be made from multimodal signals that are noisy, incomplete, or partially observable due to occlusions and blind spots. \IEEEpubidadjcol
A particularly challenging regime arises when targets are temporarily or entirely outside the camera field of view (out-of-sight), while the available localization signals (e.g., GPS/odometry or mobile sensing) are corrupted by nontrivial noise.
Despite substantial progress, 
most existing trajectory prediction methods primarily rely on visual evidence and implicitly assume reliable, consistently visible observations, or treat sensor noise as negligible. 
Consequently, they are not designed for the setting where visual cues are missing (out-of-sight) and localization signals are simultaneously noisy, which is common in safety-critical scenarios.

\begin{figure*}
\centering
\includegraphics[width=1.\linewidth]{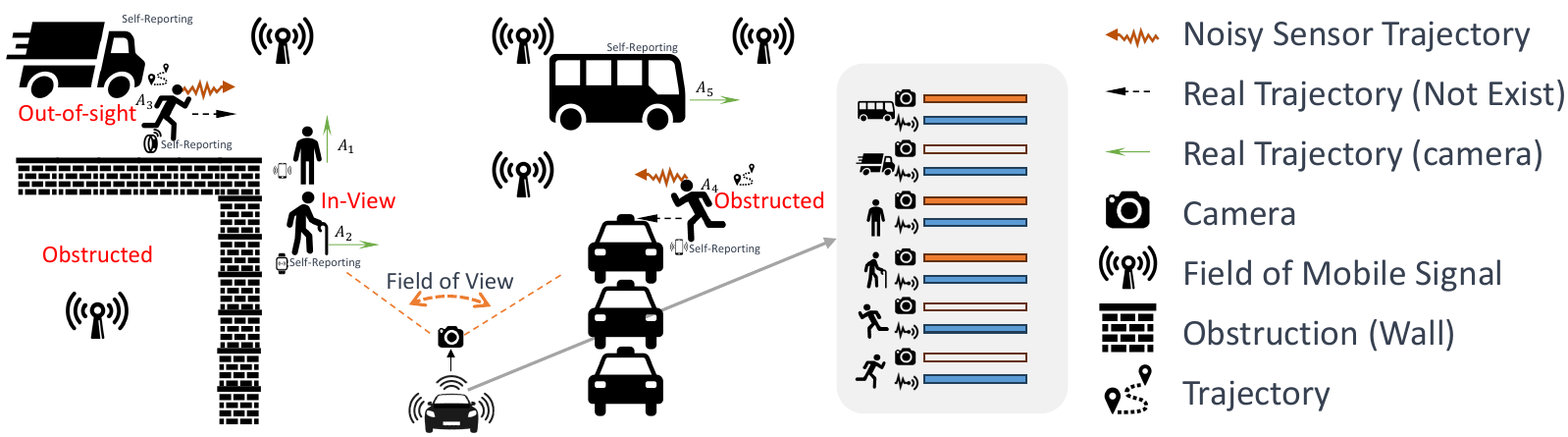}
\caption{
An illustrative example of real-world out-of-sight settings, using autonomous driving as a representative scenario. The autonomous vehicle is equipped with a camera (recording accurate visual trajectories, shown as green arrows) and a mobile signal receiver (collecting noisy sensor trajectories from IoT devices such as mobile phones, smartwatches, smart rings, or AirTags on pedestrians, as well as onboard computers or communication devices on vehicles, shown as red arrows) to monitor pedestrians and nearby vehicles. Pedestrians $A_1$ and $A_2$ are visible within the camera field of view, while $A_3$ is fully out of sight and $A_4$ is occluded by other vehicles. In addition, a bus ($A_5$) is visible in the camera view, whereas a truck ($A_6$) is moving into the vehicle's path but remains unseen because it is out of sight. Consequently, $A_3$ and $A_4$ do not have visual trajectory observations, creating substantial collision risks. The black dotted arrows denote hypothesized noise-free ground-truth trajectories that mobile sensors would ideally capture, in contrast to the observed noisy sensor trajectories (red arrows). The gray-shaded region indicates the visibility coverage of the mobile and visual modalities: white denotes no data captured, orange denotes the presence of visual trajectories, and blue denotes mobile trajectory availability.
}

\label{fig:head}
\end{figure*}

Fig.~\ref{fig:head} illustrates a representative autonomous-driving scenario. 
Due to occlusions and blind spots, some agents provide no usable visual trajectory history, while mobile/onboard localization streams remain available but noisy. 
This mismatch between missing visual evidence and corrupted localization motivates a framework that infers noise-free motion for out-of-sight agents from noisy sensor measurements.

Recent studies have begun to address incomplete observations, e.g., by imputing missing views or predicting from partially observed trajectories~\cite{xu2023uncovering,fujii2021two,zhang2023layout}. 
However, these approaches still depend on the availability of at least some visual observations. 
To the best of our knowledge, trajectory prediction for fully out-of-sight agents, where visual evidence is entirely absent, has not been systematically studied under noisy multimodal sensing.

The out-of-sight setting poses two key challenges. 
First, without direct visual supervision for the target, we must rely on localization sensors such as GPS and odometry, which are inherently noisy: GPS errors can reach 1--4 meters~\cite{huang2006low}, and odometry drift accumulates over time~\cite{zhao2017adaptive}. 
Moreover, ground-truth denoised sensor trajectories are typically unavailable, making supervised denoising impractical. 
We therefore adopt an unsupervised learning formulation to learn effective noise reduction from raw sensor streams.

Second, we aim to leverage visual information to aid denoising whenever it is available in the scene. 
For visible agents, modern tracking pipelines provide accurate image-coordinate trajectories that are effectively noise-free and can serve as a strong supervisory signal. 
To transfer this benefit to out-of-sight agents, we must establish a localization-to-vision mapping that connects sensor-space trajectories to the visual domain. 
In stationary-camera settings (e.g., surveillance), camera intrinsics are fixed and shared by all agents in the scene, enabling us to learn this mapping from the motion patterns of visible agents. 
For moving ego-camera settings, we generalize the mapping to a sequence of time-varying matrices to accommodate viewpoint changes over time.

\noindent\textbf{Contributions and extensions over the conference version~\cite{zhang2024oostraj}.}
This TPAMI extension broadens the task scope and strengthens both the method and evaluation. Specifically:
\begin{itemize}
    \item \textbf{Broader task scope.} We extend out-of-sight trajectory prediction (OST) from pedestrians to additional agent types (e.g., vehicles, pedestrians, and robots), and provide expanded evaluations on JRDB and ViFi.
    \item \textbf{Improved method.} We introduce an enhanced vision-positioning denoising framework with an improved Vision-Positioning Projection module. The method leverages accurate visual trajectories of \emph{visible} agents to learn a localization-to-vision mapping and refines noisy sensor trajectories in an unsupervised manner, without requiring ground-truth denoised supervision.
    \item \textbf{Comprehensive benchmarking and analysis.} We add extensive comparisons with classical filtering-based denoising baselines (e.g., Kalman filtering) and recent trajectory prediction models adapted to OST. We further strengthen the evaluation with (i) standard VIO metrics (ATE-RMSE, RPE$_{\text{trans}}$-RMSE, and FDE) on JRDB and ViFi, (ii) qualitative trajectory and per-frame error visualizations, and (iii) dataset-level robustness analyses via controlled multimodal sensor-noise injection that sweeps noise mode (same/cross), noise type (Gaussian, spike, drift, dropout), and noise strength. These additions are complemented by ablations, a limitations discussion, and future research directions.

\end{itemize}

\section{Related Works}
\subsection{Vision-Wireless Fusion}
The concept of vision-wireless fusion~\cite{cao2022vitag,cao2022tagging,cao2023vifit} represents an emerging interdisciplinary domain that combines the complementary strengths of visual and wireless modalities to overcome their respective limitations. The visual modality~\cite{zhangstable}, grounded in image processing and computer vision techniques, offers detailed and direct observational capabilities. However, its effectiveness is often hindered by challenges such as occlusions, constrained fields of view, high computational overhead, and its inability to handle out-of-sight scenarios. These challenges are especially pronounced in dynamic~\cite{feng2025videoorion} and complex environments, such as urban settings or industrial zones, where visual obstructions frequently occur.

On the other hand, the wireless modality provides extensive sensing and communication capabilities, utilizing technologies such as GPS, RFID, and sensor networks. These technologies excel in situations where visual data is unavailable or unreliable. Nonetheless, wireless modalities face their own set of challenges, including susceptibility to sensor noise~\cite{liu2016sensor} and environmental interference~\cite{di2002counteraction}, which can significantly degrade data quality. Furthermore, wireless data lacks the rich contextual information inherently present in visual observations, complicating tasks such as labeling and ground truth acquisition, which often depend on human interpretation aligned with visual inputs.

Despite these inherent limitations, the fusion of visual and wireless modalities has emerged as a promising avenue for addressing complex real-world challenges. Several pioneering works have demonstrated the potential of vision-wireless fusion. For instance, Liu \etal~\cite{liu2020vision} advanced the field by integrating wireless trajectories to enhance person reidentification, a task traditionally dominated by visual data. Their work addressed the complexities of human appearance variations, such as changes in clothing and posture, by leveraging the complementary strengths of wireless data. Similarly, Alahi \etal~\cite{alahi2015rgb} innovatively combined RGB-D cameras and Wi-Fi signal data to improve indoor localization, demonstrating how wireless signals from individuals' smartphones could augment visual systems in navigation tasks. In another significant contribution, Papaioannou \etal~\cite{papaioannou2015accurate} applied vision-wireless fusion in dynamic industrial environments, effectively tracking individuals across modalities to ensure robust performance in high-stakes applications.

While these foundational studies have highlighted the potential of vision-wireless fusion in various tasks, a notable gap remains in leveraging visual data for denoising noisy sensor trajectories. This limitation is particularly critical in scenarios involving out-of-sight agents, such as autonomous driving, surveillance, VLM~\cite{zhang2025unified,zhangpixels,he2025understanding}, video generation~\cite{zhang2024open}, multimodal learning~\cite{Jianglin2025,lu2025representation,li2025core}, world models~\cite{zhang2026thinkjepa}, and robotics, where sensor noise and the absence of visual context can significantly impair system reliability. Our research addresses this gap by focusing on the use of visual modalities to denoise sensor data, enabling more accurate and robust trajectory predictions. By bridging this gap, our work contributes to the advancement of vision-wireless fusion, particularly in enhancing the performance of systems operating in increasingly complex and unpredictable environments.

\subsection{Obstructed Trajectory Prediction}
Trajectory prediction is central to autonomous navigation, robotics, crowd behavior modeling, and multimodal learning~\cite{zhang2025dense,zhang2025linkedout,zhang2025vqtoken}, yet a clear gap remains between the well-studied setting of obstructed (partially observed) trajectories and the less-explored out-of-sight regime. Most prior work, such as Nikhil \etal~\cite{nikhil2018convolutional}, assumes trajectories are fully observable and therefore does not fully capture the challenges introduced by occlusions and out-of-sight targets.
To improve forecasting under partial observability, a large body of work develops richer motion and interaction representations. Social-aware approaches model interpersonal interactions and crowd dynamics~\cite{wong2024socialcircle,wong2025resonance,xie2024pedestrian,xia2022cscnet}, while multi-modal and multi-style forecasting aims to represent diverse futures~\cite{wong2023msn}. Complementary lines explore alternative generation domains, including frequency-spectrum based hierarchical modeling~\cite{wong2022view,xia2025another}, and more realistic pipelines that forecast from imperfect visual inputs, such as detection-driven forecasting~\cite{zhang2023towards}.
More recently, methods for imputing missing observations~\cite{xu2023uncovering} and fusing multimodal cues~\cite{zhang2023layout} have further advanced obstructed trajectory prediction. However, these approaches still require at least partial visual evidence or other observable cues, and thus remain insufficient when an object's trajectory is completely out of sight, a common situation in cluttered real-world environments.
In such settings, observations are not only incomplete but also noisy, leaving existing models ill-suited and creating a key research gap. We address this gap by proposing an out-of-sight trajectory prediction framework that leverages vision-positioning fusion to denoise noisy sensor signals and predict trajectories without visual observations, providing a robust solution for safety-critical autonomous systems.

\subsection{Denoising Sensor Signals}
Traditional approaches, such as the Kalman filter, have been widely adopted for state estimation in dynamic systems under noisy observation conditions. The Kalman filter is particularly effective at minimizing uncertainty by optimally combining prior state estimates with new measurements, assuming the dynamics are linear and the noise is Gaussian. However, its utility diminishes in environments characterized by complex interactions, nonlinear dynamics, and external disturbances. Scenarios involving multi-agent interactions, intricate environmental structures, and varying road geometries often introduce nonlinearities and emergent behaviors that the Kalman filter's linear framework cannot adequately model.

To address some of these shortcomings, variants like the Extended Kalman Filter (EKF) and Unscented Kalman Filter (UKF) have been developed, incorporating nonlinearity into their formulations. Yet, even these enhanced methods struggle to cope with highly dynamic systems where chaotic behaviors, such as multi-agent coordination or abrupt environmental shifts, dominate. Such scenarios frequently lead to the accumulation of noise over time, resulting in significant uncertainty amplification—a phenomenon often referred to as "Cumulative Noise" or "Butterfly Effects." These effects can propagate unpredictably, further exacerbating the difficulty of accurate trajectory estimation. To overcome these challenges, researchers have explored more sophisticated alternatives, such as particle filters, physics-based simulations, and deep learning-based models. These approaches offer greater flexibility and adaptability, allowing for more accurate modeling of complex dynamics and environmental intricacies.

Our work builds on these advancements by incorporating visual modalities into the denoising process, addressing a critical limitation of existing approaches. By leveraging noise-free visual data as a reference for correcting noisy sensor trajectories, our framework bridges the gap between traditional denoising techniques and modern data-driven methods. This integration enables more reliable predictions, particularly in scenarios where visual data is unavailable for direct observation but can be inferred through localization-vision mapping. Our vision-positioning denoising framework represents a significant step forward in addressing the challenges of trajectory prediction in noisy, dynamic environments.

\begin{figure*}[ht]
  \centering
  \vspace{20pt}
   \includegraphics[width=1.\linewidth]{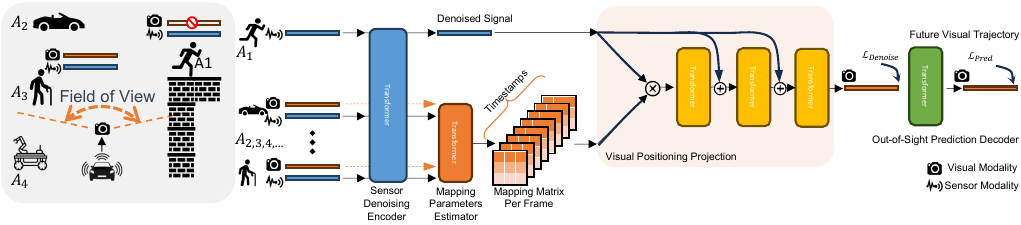}
\caption{
Overview of the Vision-Positioning Denoising and Prediction Model architecture. The figure illustrates the processing pipeline for agent trajectories, where pedestrian agent $A_1$ and autonomous robot agent $A_4$ are outside the camera field of view and can only be detected through sensor signals received by mobile receivers, while pedestrian agent $A_3$ and vehicle agent $A_2$ are observed by both the camera and sensor modality. The \textbf{Mapping Parameters Estimator Module} uses dual-modality trajectories of visible agents (e.g., $A_2$ and $A_3$) to learn a mapping matrix embedding, which is estimated for each frame separately to accommodate ego-system motion. For out-of-sight agents (e.g., $A_1$ and $A_4$), noisy mobile trajectories are refined by the \textbf{Sensor Denoising Encoder}, producing a denoised signal embedding. This embedding is then fused with the mapping matrix embedding in the \textbf{Visual Positioning Projection Module}, allowing projection into camera coordinates. The transformation is optimized with $\mathcal{L}_\text{Denoise}$. Finally, the \textbf{Out-of-Sight Prediction Decoder} takes the denoised visual signals and predicts future trajectories for agents outside the camera view, addressing the out-of-sight trajectory prediction task.
}
\label{fig:arch}
\end{figure*}

\section{Problem Definition}

\subsection{Symbol Annotations}
\label{Prob:Symbol}
Consider a set of \(N\) agents, denoted as \(A_n, n = 1, 2, \ldots, N\), observed over the time interval \([t_1:t_2]\), where \(t_s\) represents the starting timestamp, and \(t_e\) marks the endpoint. Among these agents, those visible to the camera, \(A_i\), form the set \(\mathbb{A}_{in}\), while agents outside the camera’s field of view, \(A_o\), belong to the set \(\mathbb{A}_{out}\). The sensor-derived noisy trajectory of an agent \(A_n\) over the interval \([t_1:t_2]\) is represented as \(S_{A_n}^{t_1:t_2}\). Ideally, this noisy trajectory should correspond to a noise-free localization trajectory, denoted as \(\hat{S}_{A_n}^{t_1:t_2}\). However, all agents \(A_n\) inherently possess noisy sensor trajectories \(S_{A_n}^{t_s:t_e}\), which are often plagued by significant sensor measurement noise. 

For agents within the camera’s field of view (\(A_i, i \in \mathbb{A}_{in}\)), their movements are also captured as visual trajectories \(V_{A_i}^{t_s:t_e}\) in the camera frame coordinates. These dual-modality data provide an opportunity to analyze the spatial relationship between sensor and visual trajectories, enabling more precise trajectory refinement.

\subsection{Task Definition}
Our primary objective is to tackle two interconnected challenges by leveraging the noisy sensor trajectory data \(S_{A_o}^{t_s:t_e}\) for agents that are out-of-sight:
\begin{enumerate}
    \item Develop an unsupervised denoising approach to refine the noisy sensor trajectory \(S_{A_o}^{t_s:t_e}\), mitigating the adverse effects of sensor measurement noise on prediction accuracy.
    \item Accurately predict the noise-free future visual trajectories \(V_{A_o}^{t_e:t_p}\) for out-of-sight agents \(A_o\) in the camera frame coordinates, spanning from the last observation timestamp \(t_e\) to the target prediction timestamp \(t_p\).
\end{enumerate}

It is crucial to note that out-of-sight agents (\(A_o\)) lack visual trajectory references due to their absence from the camera’s field of view. Nonetheless, through the analysis of available signals from in-view agents (\(A_i\)), a robust mapping relationship between noisy sensor trajectories and visual trajectories can be inferred, as described in Sec.~\ref{Prob:Symbol}. This allows us to address the complexities posed by environmental factors and occlusions.

Our proposed framework adopts a two-step approach: first, denoising and projecting noisy sensor trajectories into the visual modality in an unsupervised manner, leveraging signals from in-view agents as pseudo-supervision; second, employing a dedicated prediction module to generate precise, future visual trajectories for out-of-sight agents. This design ensures comprehensive handling of noisy data while enabling accurate trajectory forecasting in challenging real-world environments.

\section{Methodology}

The complete architecture of our proposed framework is shown in Fig.~\ref{fig:arch}. Our method is developed to predict noise-free future trajectories for out-of-sight agents from noisy sensor observations, without requiring direct denoising supervision or visual trajectory references.

To realize this objective, we construct an unsupervised denoising pipeline as a core part of the framework. The pipeline is designed to reduce noise in sensor measurements before it propagates to and degrades the trajectory prediction module. The formulation tackles two main challenges: how to incorporate visual cues in an indirect manner and how to build effective denoising supervision for out-of-sight agents. We break this process into several clearly defined components to provide a unified solution for noisy and incomplete sensor data.

\subsection{Sensor Denoising Encoder (SDE)}
\label{method:sde}
At the foundation of our methodology lies the Sensor Denoising Encoder (SDE), previously referred to as the Mobile Denoising Encoder. The SDE is designed to function as an unsupervised denoising module that refines noisy sensor trajectories for out-of-sight agents, even in the absence of direct supervision. Its role is pivotal in mitigating the impact of sensor noise, which can significantly degrade prediction accuracy.

The SDE's architecture comprises a Transformer model, surrounded by two fully connected layers—one preceding and one succeeding the Transformer layers. This architecture effectively captures the sequential dependencies in sensor trajectories and models the complex relationships inherent in the data.

The input to the SDE consists of noisy sensor trajectories, denoted as \(S_{A_o}^{t_s:t_e}\), where \(A_o \in \mathbb{A}_{out}\) represents an out-of-sight agent sampled from the set of out-of-sight agents. The SDE processes these trajectories and outputs refined, denoised trajectories, referred to as noise-free sensor trajectories \(\hat{S}_{A_o}^{t_s:t_e}\). The operation of the SDE can be mathematically expressed as follows:

\begin{equation}
  \hat{S}_{A_o}^{t_s:t_e}  = \mathbf{E}_{denoise} ( S_{A_o}^{t_s:t_e} ) , ~~~~~~o \sim \mathbb{A}_{out},
  \label{eq:Denoise}
\end{equation}
where \(\mathbf{E}_{denoise}\) represents the Sensor Denoising Encoder, \(\mathbb{A}_{out}\) denotes the set of out-of-sight agents, and \(A_o\) is a specific agent sampled from this set.

This denoising step plays a crucial role in ensuring that the noisy sensor data is accurately refined before being passed to subsequent modules for projection and prediction. By addressing noise at this foundational level, the SDE sets the stage for more reliable trajectory mapping and forecasting in the later stages of our framework.

However, the practical implementation of the Sensor Denoising Encoder (SDE) encounters a critical challenge: the ideal supervision in the form of noise-free real sensor trajectories, \(\hat{S}_{A_o}^{t_s:t_e}\), is inherently unavailable in real-world scenarios. The pervasive nature of sensor measurement noise complicates the possibility of obtaining such ideal data. In theory, one might consider employing high-accuracy sensors to collect data under identical conditions, which could serve as a proxy for noise-free trajectories. However, this approach presents significant practical limitations, as it is neither feasible for most datasets nor scalable across diverse application scenarios. Even under ideal conditions, higher-accuracy sensors would still introduce some level of residual measurement noise, albeit reduced, making the complete elimination of noise unattainable.

Given these constraints, constructing a truly noise-free real sensor trajectory for supervision remains beyond the capabilities of current sensor technologies. As a result, the direct use of such trajectories for supervised denoising, as suggested in Eq.~\ref{eq:Denoise}, becomes impractical. 

To address this limitation, we shift our focus towards alternative strategies for constructing effective approximations of noise-free supervision. These strategies aim to replicate the conditions of an ideal noise-free trajectory while navigating the inherent limitations of existing sensor technologies. By leveraging innovative methods that rely on indirect supervision, such as vision-based inference and unsupervised denoising approaches, we approximate the idealized conditions necessary for refining noisy sensor data. This enables the SDE to perform effectively even in challenging real-world scenarios where direct noise-free supervision is not available.

\subsection{Visual-Positioning Denoising Module (VPD)}
\label{method:vpd}
The Visual-Positioning Denoising Module (VPD) serves as the cornerstone of our denoising framework, leveraging visual information to construct noise-free supervision for the Sensor Denoising Encoder (SDE). This module addresses the intrinsic limitations of the sensor modality, which is susceptible to significant measurement noise and presents challenges for refinement. Visual data, in contrast, benefits from advancements in object detection and tracking technologies, providing highly accurate and reliable information. Moreover, the discretization inherent in image rasterization ensures that visual trajectories are naturally precise, straightforward to annotate, and effectively noise-free.

While the precision of visual trajectories makes them an ideal supervisory signal for trajectory prediction, their utility is constrained by the lack of visual data for out-of-sight agents. This limitation necessitates a robust method for translating denoised sensor trajectories into the visual domain. To accomplish this, we divide the translation process into three integral components: the Visual Positioning Projection Module (Sec.~\ref{method:vpp}), the Mapping Parameters Estimator (Sec.~\ref{method:mpe}), and the Denoising Loss (Sec.~\ref{method:Dloss}). These components collaboratively ensure the seamless mapping of sensor trajectories into the visual modality, bridging the gap between noisy sensor data and precise visual references. By integrating these elements, the VPD not only enhances the accuracy of trajectory prediction but also facilitates effective supervision in scenarios where visual data is inherently unavailable.

\subsection{Visual Positioning Projection Module (VPP)}
\label{method:vpp}
The Visual Positioning Projection Module (VPP) is a key component of our framework and is responsible for projecting noisy sensor trajectories into the visual domain. Because visual trajectories are naturally represented in the camera coordinate system, VPP targets the conversion of 3D world coordinates, derived from denoised sensor signals, into 2D camera coordinates. This projection brings sensor-based trajectories into alignment with visual trajectories by exploiting camera calibration geometry~\cite{strelow2001precise}.

At the center of VPP is the World-to-Camera transformation, which maps a 3D point \(P \in \mathbb{R}^{3 \times 1}\) in world coordinates to a 2D point \(p \in \mathbb{R}^{2 \times 1}\) in the camera plane. In practice, the height value of the 3D point is often treated as constant to simplify the 3D-to-2D projection. The transformation is formulated as follows:

\begin{equation}
  [p, 1]^T = w \cdot K \cdot R_t \cdot [P, 1]^T,
  \label{eq:Calibration}
\end{equation}

where \(K \in \mathbb{R}^{3 \times 3}\) is the camera intrinsic matrix, which contains parameters such as focal length and principal point. The scalar \(w\) is a scale factor used to maintain consistency with the camera view. \(R_t \in \mathbb{R}^{3 \times 4}\) denotes the extrinsic matrix, which includes rotation and translation terms describing the camera pose with respect to the 3D world.

For a more compact formulation, we merge these terms into a single matrix \(M \in \mathbb{R}^{3 \times 4}\), which summarizes the transformations required to project world coordinates into camera coordinates. The resulting simplified equations are:

\begin{equation}
  M = w \cdot K \cdot R_t,
  \label{eq:Matrix}
\end{equation}
\begin{equation}
  [p, 1]^T = M \cdot [P, 1]^T.
  \label{eq:seq}
\end{equation}

However, real-world scenarios necessitate considering the temporal variability of camera parameters. In stationary camera setups, such as surveillance systems, \(R_t\) remains fixed over time, simplifying the projection process. Conversely, in dynamic environments, where cameras are mounted on moving devices, \(R_t\) varies across timestamps. To account for this, we extend \(M\) into a temporal sequence, termed the Camera Matrix Embedding \(M^{t_s:t_e} \in \mathbb{R}^{(t_e-t_s) \times 3 \times 4}\), defined as:

\begin{equation}
  M^{t_s:t_e} = w \cdot K \cdot R_t^{t_s:t_e}, ~~~~~~t \in [t_s, t_e],
  \label{eq:M_seq}
\end{equation}

where \(R_t^{t_s:t_e}\) and \(M^{t_s:t_e}\) represent the extrinsic rotation and translation matrices, and the Camera Matrix Embedding over timestamps from \(t_s\) to \(t_e\), respectively.

By treating the noise-free real sensor trajectory \(\hat{S}_{A_o}^{t_s:t_e}\) as a sequence of 3D world coordinates \(P \in \mathbb{R}^{3 \times 1}\), we combine Eq.~\ref{eq:seq} and Eq.~\ref{eq:M_seq} to project \(\hat{S}_{A_o}^{t_s:t_e}\) into the visual domain:

\begin{equation}
  V_{A_o}^{t_s:t_e} = M^{t_s:t_e} \cdot \hat{S}_{A_o}^{t_s:t_e}, ~~~~~~t \in [t_s, t_e], o \sim \mathbb{A}_{out},
  \label{eq:vpp}
\end{equation}

where \(V_{A_o}^{t_s:t_e}\) represents the visual trajectory for an out-of-sight agent \(A_o\). This transformation ensures that the denoised sensor trajectory is accurately mapped into the visual modality, enabling the effective use of visual information for trajectory prediction.

\subsection{Mapping Parameters Estimator (MPE)}
\label{method:mpe}
While the Visual Positioning Projection (VPP) defined in Eq.~\ref{eq:vpp} provides a method to map denoised sensor trajectories into the visual domain, the camera matrix \(M^{t_s:t_e}\) remains inaccessible for most datasets. Typically, datasets do not provide explicit camera matrices, and they cannot be directly extracted from the images. Furthermore, in practical applications involving dynamic systems, such as autonomous driving, mobile robots, or smartphones, the camera's parameters, including its intrinsic and extrinsic properties, may change over time as the camera moves. This temporal variability significantly complicates the acquisition of the camera matrix sequence.

To address this challenge, we propose the Mapping Parameters Estimator (MPE), a module designed to predict the camera matrix embedding \(M^{t_s:t_e}\) dynamically by leveraging the correlation between visual and sensor trajectories of in-view agents. In dynamic environments, where multiple agents are simultaneously visible to the camera, both their visual trajectories (captured in the camera frame) and sensor trajectories (measured in the world frame) provide valuable clues about the underlying camera parameters. By analyzing these paired trajectories, the MPE can effectively infer the camera matrix embedding that governs the mapping relationship between the two coordinate systems.

The MPE module employs a transformer-based architecture augmented with fully connected layers positioned before and after the transformer. This design ensures the MPE's capacity to process temporal and contextual dependencies inherent in trajectory data. The prediction process is formalized as follows:

\begin{equation}
  M^{t_s:t_e} = \mathbf{E}_{mpe} ( V_{A_0}^{t_s:t_e}, S_{A_0}^{t_s:t_e}, ..., V_{A_i}^{t_s:t_e}, S_{A_i}^{t_s:t_e} ) ,~~~~~~ i \in \mathbb{A}_{in},
  \label{eq:mpe}
\end{equation}

where \(V_{A_i}^{t_s:t_e}\) and \(S_{A_i}^{t_s:t_e}\) are the visual and sensor trajectories of in-view agents \(A_i\), respectively. The set of in-sight agents is denoted by \(\mathbb{A}_{in}\). \(\mathbf{E}_{mpe}\) represents the Mapping Parameters Estimator, responsible for predicting the camera matrix embedding \(M^{t_s:t_e}\).

The MPE plays a pivotal role in dynamically estimating the camera matrix sequence across varying timestamps. By exploiting the complementary information provided by in-view agents' visual and sensor trajectories, the MPE enables robust alignment of sensor data with the visual domain, even in scenarios involving moving cameras. This capability is essential for achieving accurate trajectory prediction in dynamic, real-world environments.

\subsection{Denoising Loss}
\label{method:Dloss}
In the previous subsection, we described how to obtain the visual trajectories of out-of-sight agents using Eq.~\ref{eq:vpp}. Although ground-truth denoised sensor trajectories are unavailable, the visual modality is comparatively clean and easier to annotate or acquire, making it a suitable supervisory signal for training both the Sensor Denoising Encoder and the Mapping Parameters Estimator. By exploiting the noise-free property of visual trajectories, we build an effective supervision scheme for denoising.

The denoising loss, denoted by \(\mathcal{L}_{Denoise}\), is defined as:

\begin{equation}
  \mathcal{L}_{Denoise}  = \mathcal{L}_2 (V_{A_o}^{t_s:t_e}, \overline{V}_{A_o}^{t_s:t_e}),
  \label{eq:denoise_loss}
\end{equation}

where \(\mathcal{L}_2\) denotes the L2 loss function~\cite{buhlmann2003boosting}, which penalizes squared differences between predictions and targets. Here, \(V_{A_o}^{t_s:t_e}\) denotes the predicted visual trajectories of out-of-sight agents, and \(\overline{V}_{A_o}^{t_s:t_e}\) denotes the corresponding ground-truth visual trajectories. Notably, these ground-truth visual trajectories for out-of-sight agents are only accessible during training, which enables the model to learn both the projection mapping and denoising behavior.

\subsection{Out-of-Sight Prediction Decoder (OPD)}
\label{method:opd}
To forecast future visual trajectories for out-of-sight agents, we introduce the Out-of-Sight Prediction Decoder (OPD). Rather than predicting directly from noisy sensor trajectories, the OPD operates on the denoised visual trajectories produced by Eq.~\ref{eq:vpp}. By first projecting sensor signals into the visual domain, the decoder relies on cleaner and more structured inputs, which improves prediction accuracy and robustness.

The OPD is implemented with a transformer-based architecture due to its strong capacity for sequential modeling and temporal dependency learning. The prediction process is formulated as:

\begin{equation}
  V_{A_o}^{t_s:t_p} = \mathbf{D}_{pred} (V_{A_o}^{t_e:t_p}), ~~~~~~o \sim \mathbb{A}_{out},
  \label{eq:loss_denoise}
\end{equation}

where \(V_{A_o}^{t_e:t_p}\) denotes the observed visual trajectory of out-of-sight agent \(A_o\) over the time interval \([t_e:t_p]\), and \(V_{A_o}^{t_s:t_p}\) denotes the predicted future visual trajectory over \([t_s:t_p]\). \(\mathbf{D}_{pred}\) represents the Out-of-Sight Prediction Decoder, and \(\mathbb{A}_{out}\) denotes the set of out-of-sight agents.

To optimize OPD, we define a prediction loss \(\mathcal{L}_{Pred}\) that minimizes the gap between predicted and ground-truth trajectories:

\begin{equation}
  \mathcal{L}_{Pred}  = \mathcal{L}_2 (V_{A_o}^{t_s:t_p} , \overline{V}_{A_o}^{t_s:t_p}),
  \label{eq:loss_pred}
\end{equation}

where \(\mathcal{L}_2\) is the L2 loss, \(V_{A_o}^{t_s:t_p}\) is the predicted future visual trajectory, and \(\overline{V}_{A_o}^{t_s:t_p}\) is the corresponding ground truth. This objective encourages the predicted trajectories to remain close to the true future motion, supporting accurate and reliable out-of-sight trajectory prediction.

\subsection{Implementation Details}

During training, out-of-sight agents are randomly selected from the full set of agents in each scenario. The remaining agents that are fully observable in both the visual and noisy sensor modalities are treated as in-sight agents. These in-sight agents are used by the Mapping Parameters Estimator to infer camera-related parameters, allowing the model to obtain the camera matrix embedding required for trajectory projection.

The wireless data in our setting consists of localization signals from sensors, such as noisy GPS measurements or odometer readings, typically collected from mobile devices carried by agents. The ego-camera system (e.g., an autonomous vehicle), equipped with both a camera and a wireless receiver, records sensor trajectories together with visual trajectories of agents in the scene. The resulting multimodal data are then fed into our framework for denoising and prediction. By combining both modalities, the model can recover noise-free visual trajectories for out-of-sight agents under challenging real-world conditions.

\section{Experiments}

\begin{table*}[ht]
\centering
\begin{tabular}{c c |c c c| c c c}
\toprule
\multicolumn{2}{c}{\textbf{Dataset}}    & \multicolumn{3}{c}{\textbf{Vi-Fi Dataset}} & \multicolumn{3}{c}{\textbf{JRDB Dataset}} \\ \toprule
\textbf{Baselines}   &  \textbf{Add Module}   & \textbf{SUM$\downarrow$} & \textbf{MSE-D$\downarrow$} & \textbf{MSE-P$\downarrow$} & \textbf{SUM$\downarrow$} & \textbf{MSE-D$\downarrow$} & \textbf{MSE-P$\downarrow$}  \\ \midrule

\multirow{2}{*}{Vanilla LSTM~\cite{hochreiter1997long, shi2018lstm}}                    & + 2 Stage         & 118.44      & 38.85  & 79.59  &	81.22	&	39.93	&	41.28	\\
                                                                    & + VPD (Ours)      & 30.39       & 13.76  & 16.63  &	53.16	&	11.36	&	41.80	\\	\hline
                                                                    
\multirow{2}{*}{Vanilla RNN~\cite{rumelhart1986learning, rella2021decoder}}                & + 2 Stage         & 32.16       & 17.51  & 14.65  &	138.41	&	103.97	&	34.44	\\
                                                                    & + VPD (Ours)      & 27.78       & 13.56  & 14.22  &	31.32	&	12.31	&	19.01	\\	\hline
\multirow{2}{*}{Vanilla GRU~\cite{chung2014empirical, tran2021goal}}                    & + 2 Stage         & 40.35       & 16.88  & 23.48  &	56.56	&	17.90	&	38.66	\\
                                                                    & + VPD (Ours)      & 28.47       & 13.69  & 14.78  &	31.70	&	11.53	&	20.16	\\	\hline
\multirow{2}{*}{Vanilla Transformer~\cite{vaswani2017attention, giuliari2021transformer}} & + 2 Stage         & 28.87       & 14.22  & 14.65  &	36.99	&	14.21	&	22.79	\\
                                                                    & + VPD (Ours)      & 27.24       & 13.42  & 13.83  &	25.51	&	10.52	&	14.99	\\	\bottomrule
\end{tabular}
\caption{Plug-and-Play Experiment on the Vi-Fi and JRDB Datasets. Baselines were modified into a two-stage architecture for comparative analysis. The results highlight the effectiveness of integrating the Vision-Positioning Denoising (VPD) module, which significantly improves both denoising (MSE-D) and prediction (MSE-P) metrics across all baselines and datasets.}
\label{tab:plugandplay}
\end{table*}

\subsection{Datasets}

\subsubsection{Vi-Fi~\cite{liu2022vi} Multimodal Dataset}
The Vi-Fi dataset is a multimodal benchmark specifically constructed for vision-wireless systems, with an emphasis on associating pedestrian identities across visual and wireless modalities. In this dataset, wireless signals are collected from smartphones carried by pedestrians, including Fine Time Measurement (FTM), Inertial Measurement Unit (IMU), and noisy GPS signals. At the same time, an RGB-D camera surveillance setup, together with a wireless receiver, is deployed in either wall-mounted or bicycle-mounted configurations to record pedestrian bounding boxes, enabling synchronized collection of visual and wireless data. Importantly, the wireless signals in Vi-Fi contain substantial noise, which must be handled during model training and evaluation.

The dataset contains 90 sequences, each with a duration of about 3 minutes, spanning both indoor and outdoor scenes. Indoor sequences were collected with five authorized participants and no passersby, while outdoor sequences include three active participants and 12 passersby. In our experiments, we use the noisy GPS signals as noisy mobile trajectories and use visual points as the visual modality. To construct out-of-sight cases, we artificially mask one pedestrian in each sequence, creating a controlled setting for evaluating our method.

\subsubsection{JackRabbot~\cite{martin2021jrdb} Dataset (JRDB)}
The JRDB dataset, collected using the JackRabbot social mobile robot, is designed for research on autonomous robot navigation and social robotics in human-centered environments. It provides 60,000 annotated frames, including 2.4 million 2D human bounding boxes and 2.8 million 3D human bounding boxes, captured with a combination of 360$^\circ$ RGB cameras (five cameras) and LIDAR point clouds. The dataset includes both occluded and visible pedestrians. Since each individual camera covers only a 72-degree field of view, out-of-sight situations naturally occur among agents.

The 2D and 3D bounding boxes are manually labeled but have relatively low temporal resolution, so we upsample them with linear interpolation to obtain frame-level continuity. Although this interpolation introduces additional noise, it provides a practical basis for trajectory estimation. Occluded pedestrians in the LIDAR modality are manually estimated in terms of size and position. In our experiments, we use the center points of the 3D bounding boxes as noisy sensor trajectories and use visual points as visual trajectories. To simulate out-of-sight conditions, we randomly mask one pedestrian and use the visual trajectories of the remaining visible pedestrians to estimate camera parameters through our model.

\subsection{Evaluation Setup}
For both JRDB and Vi-Fi, we observe the first 100 timestamps of out-of-sight sensor trajectories and predict the following 100 timestamps in the visual modality. In parallel, we use paired in-sight visual trajectories and noisy sensor trajectories over 100 timestamps to estimate camera matrix embeddings via the Mapping Parameters Estimator. This protocol enables joint evaluation of the denoising and prediction modules under realistic conditions.

\subsection{Metrics}
As the first framework that denoises out-of-sight sensor trajectories and projects them into the visual domain for noise-free trajectory prediction, we adopt tailored metrics to evaluate performance comprehensively. Specifically, we use Mean Square Error per Timestamp (\textbf{MSE-T}) from~\cite{zhang2023layout}, which measures the average pixel-distance error across time.

To assess denoising and prediction quality separately, we compute MSE-T on the projected out-of-sight visual trajectories and on the predicted visual trajectories, yielding \textbf{MSE-D} (denoising MSE) and \textbf{MSE-P} (prediction MSE), respectively. We further define a combined metric, \textbf{SUM}, which adds MSE-D and MSE-P to provide an overall measure of the denoising--prediction trade-off.

\subsection{Baselines}
To evaluate our method for noisy sensor trajectory denoising and out-of-sight trajectory prediction, we include a diverse set of baselines spanning classical sequence models and stronger recent approaches. The compared methods are:

\begin{itemize}
    \item \textbf{Vanilla LSTM}~\cite{shi2018lstm, hochreiter1997long}: A long short-term memory network, commonly used for sequence modeling because of its ability to capture long-range temporal dependencies.
    \item \textbf{Vanilla RNN}~\cite{rella2021decoder, rumelhart1986learning}: A standard recurrent neural network that serves as a foundational sequence model but is susceptible to vanishing gradients.
    \item \textbf{Vanilla GRU}~\cite{tran2021goal, chung2014empirical}: A gated recurrent unit model that provides a lighter alternative to LSTM while maintaining strong performance in trajectory forecasting tasks.
    \item \textbf{Vanilla Transformer}~\cite{vaswani2017attention, giuliari2021transformer}: A self-attention-based architecture widely adopted for sequence modeling due to its parallel computation and long-range dependency modeling capability.
    \item \textbf{ViTag}~\cite{liu2021lost, cao2022vitag}: A state-of-the-art method on the Vi-Fi dataset originally developed for vision-wireless association. We adapt it to match the output format and requirements of our task.
\end{itemize}

\begin{table*}[ht]
\centering
\begin{tabular}{c |c c c| c c c}
\toprule
\textbf{Dataset}    & \multicolumn{3}{c}{\textbf{Vi-Fi Dataset}} & \multicolumn{3}{c}{\textbf{JRDB Dataset}} \\ \toprule
\textbf{Module Components}   & \textbf{SUM$\downarrow$} & \textbf{MSE-D$\downarrow$} & \textbf{MSE-P$\downarrow$} & \textbf{SUM$\downarrow$} & \textbf{MSE-D$\downarrow$} & \textbf{MSE-P$\downarrow$}  \\ \midrule
w/o SDE in Sec.~\ref{method:sde}                & 32.12 & 17.01 & 15.12 & 29.66 & 14.35 & 15.31 \\			 	 	 
w/o MPE in Sec.~\ref{method:mpe}                & 32.33 & 17.65 & 14.68 & 32.94 & 17.90 & 15.03 \\	
w/o VPP in Sec.~\ref{method:vpp}                & 28.42 & 14.05 & 14.37 & 32.48 & 14.45 & 18.03 \\
w/o OPD in Sec.~\ref{method:opd}                & 27.33 & 13.65 & 13.68 & 30.01 & 14.99 & 15.02 \\	
\midrule
Ours (CVPR Version)~\cite{zhang2024oostraj}    & 27.24 & 13.42 & 13.83 & 25.51 & 10.52 & 14.99 \\ 
Ours (Journal Version)                         & 23.09 & 11.86 & 11.23 & 21.97 & 10.84 & 11.13 \\
\bottomrule
\end{tabular}
\caption{Ablation study of module components, showing the effect of each component on performance on the Vi-Fi and JRDB datasets. We report MSE-D for denoising quality, MSE-P for prediction quality, and SUM as the combined performance metric.}
\label{tab:ablation}
\end{table*}

\subsection{Ablation Study}
We perform an ablation study to measure the contribution of each module in our model. By removing one component at a time, we analyze its effect on overall performance. The quantitative results are reported in Table~\ref{tab:ablation}, which demonstrates the importance of each module to the final performance.

Removing the Camera Parameter Estimator (CPE) leads to a clear performance drop. Without CPE, the model cannot estimate the camera matrix sequence required to build the visual-positioning relationship, and the framework degenerates into a simpler encoder-decoder design. As a result, a key mechanism for connecting sensor and visual modalities is lost. Likewise, removing the Mobile Denoising Encoder (MDE) causes an even larger degradation. Without MDE, sensor noise is insufficiently suppressed and propagates to later stages, including the Visual Positioning Projection (VPP) module.

The VPP module is also essential because it explicitly uses geometric camera calibration to project noisy sensor trajectories into the visual domain. When VPP is removed, the model’s ability to accurately project out-of-sight trajectories is significantly weakened, confirming the importance of geometric priors in the framework. In addition, removing the Out-of-Sight Prediction Decoder (OPD) harms trajectory forecasting performance. Without OPD, the model must jointly handle denoising and future prediction in a less specialized way, which reduces prediction quality.

Overall, the ablation results verify the necessity of all modules in the proposed architecture. The Camera Parameter Estimator, Mobile Denoising Encoder, Visual Positioning Projection Module, and Out-of-Sight Prediction Decoder work together to achieve the strong performance of our model on noise-free out-of-sight trajectory prediction. Each component addresses a different part of the problem, and their combination improves the robustness and effectiveness of the full framework.

\subsection{Quantitative Comparison with Recent Trajectory Prediction Approaches}

\subsubsection{Baselines}
To evaluate the effectiveness of our proposed method, we compare it with recent state-of-the-art (SOTA) trajectory prediction approaches. Because our task jointly addresses noisy GPS denoising and visual trajectory prediction for out-of-sight agents, direct comparison is nontrivial due to the absence of closely matched prior methods. For a fair and meaningful evaluation, we choose HiVT~\cite{zhou2022hivt} and AutoBots~\cite{girgis2022latent} as baselines. We adapt these methods to our setting by modifying their input-output interfaces, mapping modules, and other task-incompatible components. With these changes, the baselines can take noisy sensor trajectories of out-of-sight agents as input and use trajectory pairs of in-sight agents as contextual information.

\subsubsection{Results Analysis}
The quantitative comparison is reported in Table~\ref{table:TP}. Our model consistently surpasses the baseline methods on all three evaluation metrics. In particular, our approach shows a clear advantage in denoising, which reflects the effectiveness of the Vision-Positioning design in incorporating environmental information through camera-parameter estimation. The gain in denoising quality further leads to improved trajectory prediction performance, indicating a strong connection between these two stages.

We also observe that baseline methods with stronger denoising results tend to produce better trajectory prediction accuracy. This pattern is consistent with the observations from our plug-and-play experiment (see Table~\textcolor{red}{2} in the main paper). Overall, these results highlight the importance of robust denoising for out-of-sight trajectory prediction. By suppressing sensor noise and leveraging environmental context, our method provides stronger performance for challenging noisy and out-of-sight scenarios.

\begin{table*}[t]
\centering
\begin{tabular}{c |c c c| c c c}
\toprule
\textbf{Dataset}    & \multicolumn{3}{c}{\textbf{Vi-Fi Dataset}} & \multicolumn{3}{c}{\textbf{JRDB Dataset}} \\ \toprule
\textbf{Baselines}   & \textbf{SUM$\downarrow$} & \textbf{MSE-D$\downarrow$} & \textbf{MSE-P$\downarrow$} & \textbf{SUM$\downarrow$} & \textbf{MSE-D$\downarrow$} & \textbf{MSE-P$\downarrow$}  \\ \midrule
ViTag~\cite{liu2021lost}                            & 200.90 & 100.53 & 100.37 & 143.08 & 71.23 & 71.85 \\	
Vanilla LSTM~\cite{hochreiter1997long, shi2018lstm}                     & 116.01 & 58.31 & 57.70 & 56.05 & 27.98 & 28.07 \\	
Vanilla GRU~\cite{chung2014empirical, tran2021goal}                     & 57.34 & 28.69 & 28.65 & 71.91 & 35.83 & 36.07 \\	
Vanilla RNN~\cite{rumelhart1986learning, rella2021decoder}                 & 31.61 & 15.92 & 15.69 & 112.40 & 56.00 & 56.41 \\	
Vanilla Transformer~\cite{vaswani2017attention, giuliari2021transformer}  & 28.33 & 14.26 & 14.08 & 33.37 & 16.71 & 16.66 \\	
\midrule
Ours (CVPR Version)~\cite{zhang2024oostraj} & 27.24 & 13.42 & 13.83 & 25.51 & 10.52 & 14.99 \\
Ours (Journal Version) & 23.09 & 11.86 & 11.23 & 21.97 & 10.84 & 11.13 \\
\bottomrule
\end{tabular}
\caption{Quantitative comparison of models on the Vi-Fi and JRDB datasets. Metrics include SUM (combined error), MSE-D (denoising error), and MSE-P (prediction error). Our method, in both its CVPR and Journal versions, demonstrates superior performance compared to baselines, particularly in denoising noisy sensor trajectories and predicting visual trajectories for out-of-sight agents.}
\label{tab:quan}
\end{table*}

\subsection{Quantitative Comparison Experiments}
We conduct quantitative experiments to compare our method with five baseline models on two datasets (Vi-Fi and JRDB) using three metrics (SUM, MSE-D, and MSE-P), as reported in Table~\ref{tab:quan}. The results show that our model outperforms all baselines. In particular, although ViTag is a state-of-the-art (SOTA) method on Vi-Fi for visual-wireless identity association, its performance drops notably on our denoising and prediction tasks. This result highlights the distinct challenges of our setting, which requires denoising noisy sensor trajectories and projecting out-of-sight agents into the visual domain.

A key observation is that our model shows a larger margin in denoising than in prediction. This reflects the difficulty of cleaning noisy sensor trajectories and projecting them accurately into visual coordinates. We also observe a strong positive correlation between denoising and prediction quality across baselines: better denoising generally leads to better prediction. This supports our design choice of treating sensor denoising as a primary step for accurate out-of-sight visual trajectory prediction.

We further note that baseline behavior varies across datasets. For example, Vanilla RNN performs relatively well on Vi-Fi but degrades on JRDB, likely because the two datasets have different sensor-noise characteristics. In contrast, our vision-positioning denoising model, which is designed to handle diverse noise conditions with visual supervision, maintains stable performance on both datasets. We investigate this point further in the Plug-and-Play experiments.

\subsection{Plug-and-Play Experiments}
In the Plug-and-Play experiment (Table~\ref{tab:plugandplay}), we test whether our vision-positioning denoising (VPD) model can improve the denoising and prediction performance of baseline models. For a fair comparison, each baseline is reformulated into a two-stage pipeline, with a denoising stage followed by a prediction stage. The ``+ VPD'' rows indicate that our VPD module (Sec.~\ref{method:vpd}) is inserted into the corresponding baseline, allowing us to measure its contribution directly.

Table~\ref{tab:plugandplay} shows that adding VPD consistently improves all baselines on both denoising and prediction metrics. For instance, Vanilla RNN, which is more sensitive to sensor noise on JRDB, gains clear denoising improvements after integrating VPD, and these gains also lead to better prediction accuracy. This behavior demonstrates that VPD effectively refines noisy sensor trajectories and provides cleaner inputs for downstream trajectory forecasting.

Overall, the Plug-and-Play results confirm that VPD is a flexible and effective module that can strengthen existing models in both denoising and prediction. By reducing sensor noise and improving trajectory quality before forecasting, our approach offers a practical enhancement for out-of-sight trajectory prediction under noisy conditions.

\begin{table*}[t]
\centering
\begin{tabular}{c |c c c| c c c}
\toprule
\textbf{Dataset}    & \multicolumn{3}{c}{\textbf{Vi-Fi Dataset}} & \multicolumn{3}{c}{\textbf{JRDB Dataset}} \\ \toprule
\textbf{Baselines}   & \textbf{SUM$\downarrow$} & \textbf{MSE-D$\downarrow$} & \textbf{MSE-P$\downarrow$} & \textbf{SUM$\downarrow$} & \textbf{MSE-D$\downarrow$} & \textbf{MSE-P$\downarrow$}  \\ \midrule
HIVT~\cite{zhou2022hivt}                      & 33.48  &  14.80 &  18.68 &	29.19	&	13.55	&	15.64 \\	
AutoBots~\cite{girgis2022latent}                  & 35.78 & 16.14  & 19.63 &	37.97	&	19.17	&	18.80 \\	
\midrule
Ours (CVPR Version)~\cite{zhang2024oostraj} & 27.24 & 13.42 & 13.83 &	25.51	&	10.52	&	14.99 \\	
Ours (Journal Version) & 23.09 & 11.86 & 11.23 & 21.97 & 10.84 & 11.13 \\
\bottomrule
\end{tabular}
\caption{Quantitative comparison results with recent state-of-the-art (SOTA) methods in trajectory prediction. Our method demonstrates superior performance across all metrics, particularly in denoising noisy sensor trajectories and predicting visual trajectories of out-of-sight agents.}
\label{table:TP}
\end{table*}

\begin{table}[thb]
\centering
\begin{tabular}{c |c c c}
\toprule
\textbf{Baselines} & \textbf{SUM$\downarrow$} & \textbf{MSE-D$\downarrow$} & \textbf{MSE-P$\downarrow$} \\ \midrule
Kalman Filter~\cite{kalman1960new} + Transformer               & 35.18     & 17.63  & 17.54 \\	
VPD + Transformer (Ours)                  & 27.24     & 13.42  & 13.83 	\\
\bottomrule
\end{tabular}
\caption{Comparison with traditional denoising approaches. Our VPD module significantly outperforms traditional Kalman filter-based methods, showcasing its ability to handle complex noise and environmental dynamics effectively.}
\label{table:trad}
\end{table}

\begin{figure}[tb]
  \centering
  \vspace{20pt}
   \includegraphics[width=1.\linewidth]{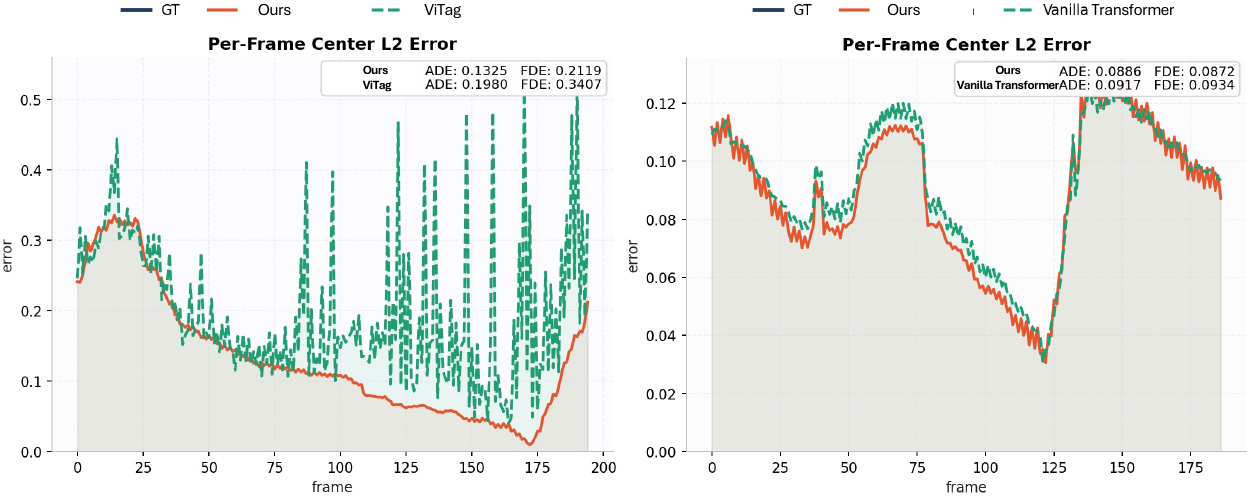}
    \caption{Per-frame center-point $\ell_2$ error on representative test sequences. 
    We plot $\|\hat{\mathbf{c}}_t-\mathbf{c}_t\|_2$ over time for ground truth (GT) and different predictors. 
    Left: Ours vs.\ ViTag. Right: Ours vs.\ Vanilla Transformer. 
    The inset reports ADE and FDE computed on valid frames after robust filtering.}
\label{fig:vis}
\end{figure}

\subsection{Trajectory and Error Visualization.}
As shown in Fig.~\ref{fig:vis}, we visualize per-frame trajectory error using the $\ell_2$ distance between predicted and ground-truth center points. 
For each test sample, we load the saved prediction sequences (our VisionPosition outputs and baseline outputs) together with the corresponding ground truth (GT), and apply robust frame filtering to exclude invalid or empty frames. 
We then compute and plot the per-frame center-point error $\|\hat{\mathbf{c}}_t-\mathbf{c}_t\|_2$, and report summary statistics (ADE/FDE) computed over valid frames only. 
Overall, our method achieves lower error across most time steps and exhibits a smoother error profile with fewer abrupt spikes than both ViTag and the Vanilla Transformer. 
Against the Vanilla Transformer, we observe consistent but smaller improvements, indicating improved stability under comparable error regimes.

\subsection{Dataset-level noise prevalence and robustness evaluation.}
As summarized in Table.~\ref{tab:noise-agg}, we conduct dataset-level robustness analyses on \textbf{ViFi} and \textbf{JRDB} by injecting controlled synthetic noise into multimodal sensor streams at evaluation time. 
We sweep over (i) \emph{noise mode} (\texttt{same} vs.\ \texttt{cross}), (ii) \emph{noise type} (\texttt{gaussian}, \texttt{spike}, \texttt{drift}, \texttt{dropout}), and (iii) \emph{noise strength} (default $\{0, 0.01, 0.03, 0.05, 0.1\}$). 
In \texttt{same} mode, all sensor modalities share the same noise type and strength, which evaluates robustness under homogeneous corruption. 
In \texttt{cross} mode, different modalities are corrupted with different noise types according to a predefined (or configurable) profile; we denote this mixed configuration as \texttt{cross\_mixed}. 
For each run, we parse logs to extract loss components and standard VIO metrics, and report both per-run results and aggregated statistics.

Specifically, we group results by \texttt{dataset}, \texttt{model}, \texttt{noise\_mode} (\texttt{same}/\texttt{cross}), \texttt{noise\_type} (a single type in \texttt{same}, and \texttt{cross\_mixed} in \texttt{cross}), and \texttt{noise\_strength}. 
Here, \texttt{MSE-D} measures denoising/reconstruction error and \texttt{MSE-P} measures prediction error. 
We additionally report standard VIO metrics: \texttt{ATE-RMSE}, \texttt{RPE$_{\text{trans}}$-RMSE}, and \texttt{FDE}. 
Overall, our model exhibits substantially smaller performance degradation under increasing noise levels compared to the baselines, indicating stronger robustness to multimodal sensor corruption.

\subsection{Standard VIO metrics.}
We adapt standard visual-inertial odometry (VIO) metrics to better suit the out-of-sight trajectory prediction setting. 
Let $\hat{\mathbf{p}}_{b,t}\in\mathbb{R}^{d}$ and $\mathbf{p}_{b,t}\in\mathbb{R}^{d}$ denote the predicted and ground-truth translation vectors for batch index $b$ and time step $t$, respectively, where we use the first \texttt{pos\_dim} dimensions (e.g., $d=3$ when available). 
We define the per-frame translation error as
\begin{equation}
e_{b,t} = \left\|\hat{\mathbf{p}}_{b,t}-\mathbf{p}_{b,t}\right\|_{2}.
\end{equation}
We compute \textbf{ATE-RMSE} as
\begin{equation}
\mathrm{ATE}_{\mathrm{RMSE}}=\sqrt{\frac{1}{BT}\sum_{b=1}^{B}\sum_{t=1}^{T} e_{b,t}^{2}},
\end{equation}
\textbf{FDE} (final displacement error) as
\begin{equation}
\mathrm{FDE}=\frac{1}{B}\sum_{b=1}^{B}\left\|\hat{\mathbf{p}}_{b,T}-\mathbf{p}_{b,T}\right\|_{2},
\end{equation}
and \textbf{RPE(trans)-RMSE} with step size $\delta$ (default $\delta=1$) by comparing $\delta$-step increments:
\begin{equation}
\Delta\hat{\mathbf{p}}_{b,t}=\hat{\mathbf{p}}_{b,t+\delta}-\hat{\mathbf{p}}_{b,t},\quad
\Delta\mathbf{p}_{b,t}=\mathbf{p}_{b,t+\delta}-\mathbf{p}_{b,t},
\end{equation}
\begin{equation}
\mathrm{RPE}_{\mathrm{trans,RMSE}}=\sqrt{\frac{1}{B(T-\delta)}\sum_{b=1}^{B}\sum_{t=1}^{T-\delta}\left\|\Delta\hat{\mathbf{p}}_{b,t}-\Delta\mathbf{p}_{b,t}\right\|_{2}^{2}}.
\end{equation}
If $T\le \delta$, RPE is undefined and we report it as NaN. 
As shown in Table.~\ref{tab:test-metrics}, these standard VIO metrics provide an additional and widely adopted view of performance, further validating the advantage of our method over the baselines.

\begin{table*}[t]
\centering
\small
\setlength{\tabcolsep}{5pt}
\renewcommand{\arraystretch}{1.15}
\begin{tabular}{l rrr rrr}
\toprule
& \multicolumn{3}{c}{\textbf{JRDB}} & \multicolumn{3}{c}{\textbf{ViFi}} \\
\cmidrule(lr){2-4}\cmidrule(lr){5-7}
\textbf{Model} 
& \textbf{ATE-RMSE}$\downarrow$ & \textbf{RPE$_{\text{trans}}$-RMSE}$\downarrow$ & \textbf{FDE}$\downarrow$
& \textbf{ATE-RMSE}$\downarrow$ & \textbf{RPE$_{\text{trans}}$-RMSE}$\downarrow$ & \textbf{FDE}$\downarrow$ \\
\midrule
Vanilla ViTag
& 27.50 & 31.84 & 35.40
& 20.15 & 13.70 & 17.16 \\
Vanilla LSTM
& 13.85 &  6.22 & 12.15
& 16.15 &  4.52 & 20.06 \\
Vanilla Transformer
& 13.21 &  3.64 & 12.01
& 15.99 &  4.27 & 13.79 \\
\midrule
\textbf{Ours}
& \textbf{13.06} & \textbf{4.23} & \textbf{11.77}
& \textbf{15.89} & \textbf{4.26} & \textbf{13.63} \\
\bottomrule
\end{tabular}
\caption{Test results on JRDB and ViFi using standard VIO metrics: absolute trajectory error (ATE-RMSE), translational relative pose error (RPE$_{\text{trans}}$-RMSE), and final displacement error (FDE). Lower is better.}
\label{tab:test-metrics}
\end{table*}

\begin{table*}[t]
\centering
\small
\setlength{\tabcolsep}{4pt}
\renewcommand{\arraystretch}{1.15}
\begin{tabular}{lll r r r r r}
\toprule
\multicolumn{3}{c}{\textbf{Noise Configuration}} &
\multicolumn{2}{c}{\textbf{Distance Error}} &
\multicolumn{3}{c}{\textbf{VIO Error}} \\
\cmidrule(lr){1-3}\cmidrule(lr){4-5}\cmidrule(lr){6-8}
\textbf{Mode} & \textbf{Type} & \textbf{Strength} &
\textbf{MSE-D}$\downarrow$ & \textbf{MSE-P}$\downarrow$ &
\textbf{ATE-RMSE}$\downarrow$ & \textbf{RPE$_{\text{trans}}$-RMSE}$\downarrow$ & \textbf{FDE}$\downarrow$ \\
\midrule
cross & cross\_mixed & 0    & 12.708770 & 12.160218 & 14.580712 & 4.2792170 & 12.615881 \\
cross & cross\_mixed & 0.01 & 12.708770 & 12.160218 & 14.580712 & 4.2792170 & 12.615881 \\
cross & cross\_mixed & 0.03 & 12.708770 & 12.160218 & 14.580712 & 4.2792170 & 12.615881 \\
cross & cross\_mixed & 0.05 & 12.708770 & 12.160218 & 14.580712 & 4.2792170 & 12.615881 \\
cross & cross\_mixed & 0.1  & 12.708770 & 12.160218 & 14.580712 & 4.2792170 & 12.615881 \\
\midrule
same  & drift    & 0    & 12.708760 & 12.160218 & 14.580712 & 4.2792190 & 12.615879 \\
same  & drift    & 0.01 & 12.708759 & 12.160219 & 14.580712 & 4.2792225 & 12.615881 \\
same  & drift    & 0.03 & 12.708758 & 12.160218 & 14.580713 & 4.2792305 & 12.615883 \\
same  & drift    & 0.05 & 12.708756 & 12.160218 & 14.580713 & 4.2792395 & 12.615884 \\
same  & drift    & 0.1  & 12.708755 & 12.160217 & 14.580713 & 4.2792615 & 12.615891 \\
\midrule
same  & dropout  & 0    & 12.708760 & 12.160218 & 14.580712 & 4.2792190 & 12.615879 \\
same  & dropout  & 0.01 & 12.708761 & 12.160218 & 14.580712 & 4.2792180 & 12.615880 \\
same  & dropout  & 0.03 & 12.708761 & 12.160219 & 14.580712 & 4.2792160 & 12.615878 \\
same  & dropout  & 0.05 & 12.708761 & 12.160219 & 14.580712 & 4.2792135 & 12.615879 \\
same  & dropout  & 0.1  & 12.708762 & 12.160218 & 14.580712 & 4.2792065 & 12.615877 \\
\midrule
same  & gaussian & 0    & 12.708760 & 12.160218 & 14.580712 & 4.2792190 & 12.615879 \\
same  & gaussian & 0.01 & 12.708760 & 12.160219 & 14.580712 & 4.2792190 & 12.615881 \\
same  & gaussian & 0.03 & 12.708761 & 12.160218 & 14.580712 & 4.2792190 & 12.615883 \\
same  & gaussian & 0.05 & 12.708764 & 12.160218 & 14.580712 & 4.2792210 & 12.615886 \\
same  & gaussian & 0.1  & 12.708777 & 12.160219 & 14.580713 & 4.2792275 & 12.615892 \\
\midrule
same  & spike    & 0    & 12.708760 & 12.160218 & 14.580712 & 4.2792190 & 12.615879 \\
same  & spike    & 0.01 & 12.708760 & 12.160218 & 14.580712 & 4.2792205 & 12.615885 \\
same  & spike    & 0.03 & 12.708771 & 12.160218 & 14.580713 & 4.2792300 & 12.615898 \\
same  & spike    & 0.05 & 12.708794 & 12.160220 & 14.580715 & 4.2792445 & 12.615910 \\
same  & spike    & 0.1  & 12.708869 & 12.160228 & 14.580724 & 4.2793070 & 12.615953 \\
\bottomrule
\end{tabular}
\caption{Aggregated robustness results (mean over two successful runs) under controlled synthetic noise injection at evaluation time. We vary noise mode (\texttt{same} vs.\ \texttt{cross}), noise type, and noise strength. \textbf{MSE-D} and \textbf{MSE-P} denote denoising/reconstruction and prediction errors, respectively; VIO metrics include ATE-RMSE, RPE$_{\text{trans}}$-RMSE, and FDE (lower is better).}
\label{tab:noise-agg}
\end{table*}

\subsection{Comparison with Traditional Denoising Approaches}

\subsubsection{Baseline}
Denoising sensor signals has long been a focus of research, with traditional methods like the Kalman filter~\cite{kalman1960new} being widely adopted for this purpose. The Kalman filter's popularity stems from its ability to estimate the state of a dynamic system under noisy observations by assuming linear dynamics and Gaussian noise. However, the application of deep learning-based methods for sensor signal denoising has gained traction only in recent years. A significant barrier to adopting supervised learning in this context is the lack of ground truth data. Since sensor trajectories are captured in real-world environments, dataset collectors cannot fully eliminate noise to create noise-free signals as ground truth. This limitation renders supervised learning infeasible and reinforces the reliance on traditional methods like the Kalman filter as preprocessing tools in signal processing pipelines. To evaluate the effectiveness of our approach, we select the Kalman filter as a baseline for comparison in this experiment.

\subsubsection{Results Analysis}
Traditional methods such as the Kalman filter operate under simplifying assumptions, including linear system dynamics and Gaussian noise, which restrict their applicability in capturing the complex interactions and non-linear behaviors inherent in real-world autonomous systems. These methods focus solely on denoising based on state transitions without accounting for dynamic environmental changes or the intricate relationships between agents in autonomous systems.

As shown in Table~\ref{table:trad}, our proposed Vision-Positioning Denoising (VPD) model significantly outperforms the combination of the Kalman filter for denoising and a transformer model for trajectory prediction. Specifically, the VPD model demonstrates superior performance across both denoising and prediction metrics. This improvement highlights the VPD model's ability to transcend the limitations of traditional methods by leveraging advanced geometric calibration and context-aware data fusion, enabling it to model complex environmental dynamics and multi-agent interactions effectively. These results emphasize the transformative potential of our approach in addressing the limitations of traditional denoising techniques for out-of-sight trajectory prediction tasks.

\section{Limitation Discussion}
\subsection{Implicit Constraints in Camera Calibration}
The maintaining calibration constraints is important. In many practical scenarios, intrinsic matrices are not included in datasets, and real-world camera operations such as zooming and autofocus can modify it. Furthermore, collecting extrinsic matrices is arduous and error-prone, as it involves solving PnP problems with noisy high-precision GPS data\cite{liu2023vifi}. Considering the dynamic nature of cameras in autonomous vehicles and robotics, our method's capability to estimate both matrices as a unified embedding allows for robustness in these more complex, realistic scenarios. When intrinsic matrices are available, our approach can seamlessly integrate them by directly multiplying the provided camera matrix embeddings, ensuring compatibility with traditional calibration methods.

\subsection{The Perception Distance of Out-of-Sight}
Since the out-of-sight trajectory prediction model can involve agents very far from the objects, We need to consider the situation like what if multimoadl modality receive a very long distance signal. And furthermore, what is the most far distance can our model receive? Like if we receive a mobile signal from a few miles away, the performance of our model will definetly dropped.
Our model does not necessitate that in-sight and out-of-sight objects be at the same distance.
As both share the same camera projection matrix and the trajectories are captured at various distances over different timestamps, a strict distance parity is not essential. However, model generalization is contingent upon the data distribution; objects positioned at extreme distances could lead to out-of-distribution issues, potentially affecting performance. Currently, our model effectively covers distances within our dataset's scope, and the performance at significantly farther distances, like several miles away, remains to be evaluated.
\section{Conclusion}
This study addresses an important limitation in trajectory prediction research and practical deployments: many existing settings implicitly assume that every agent has visual observations, which does not hold for fully out-of-sight agents. We introduce a new task that predicts visual trajectories of out-of-sight agents from noisy sensor trajectories. To solve this task, we propose a Visual-Positioning Denoising Module, which handles the absence of direct visual references by using camera calibration to estimate camera matrix sequences and build a vision-positioning mapping. We also address the challenge that noisy sensor trajectories are heavily corrupted and lack direct denoising supervision by constructing supervision through visual-positioning projection.

Extensive experiments show that our pipeline achieves strong performance in both denoising and out-of-sight trajectory prediction. The results demonstrate that incorporating visual-positioning information into the denoising process leads to better downstream trajectory prediction. To the best of our knowledge, this is the first work to predict trajectories of completely out-of-sight agents from noisy sensor trajectories, and the first to estimate vision-positioning projections for denoising out-of-sight noisy sensor data.

The methods and findings in this work support future research on trajectory prediction in more realistic settings, especially when visual observations are incomplete or sensor inputs are noisy. By introducing a unified framework that combines visual and sensor modalities, our study provides a practical basis for future efforts to improve the safety, reliability, and accuracy of autonomous systems in dynamic real-world environments.


\ifCLASSOPTIONcaptionsoff
  \newpage
\fi

    

{
\bibliographystyle{ieee_fullname}
\bibliography{main}

@String(CVPR= {IEEE Conf. Comput. Vis. Pattern Recog.})

@String(ECCV= {Eur. Conf. Comput. Vis.})

@String(ICPR = {Int. Conf. Pattern Recog.})

@String(CVPR  = {CVPR})

@String(ECCV  = {ECCV})

@String(ICPR  = {ICPR})

@article{ramesh2021dalle,
  author       = {Aditya Ramesh and Mikhail Pavlov and Gabriel Goh and Scott Gray and Mark Chen and Rewon Child and Vedant Misra and Pamela Mishkin and Gretchen Krueger and Sandhini Agarwal and Ilya Sutskever},
  title        = {DALL·E: Creating Images from Text},
  year         = {2021},
  journal = {\url{https://openai.com/dall-e}}
}

@inproceedings{zhang2024oostraj,
  title={OOSTraj: Out-of-Sight Trajectory Prediction With Vision-Positioning Denoising},
  author={Zhang, Haichao and Xu, Yi and Lu, Hongsheng and Shimizu, Takayuki and Fu, Yun},
  booktitle={Proceedings of the IEEE/CVF Conference on Computer Vision and Pattern Recognition},
  pages={14802--14811},
  year={2024}
}

@inproceedings{cao2022vitag,
  title={ViTag: Online WiFi Fine Time Measurements Aided Vision-Motion Identity Association in Multi-person Environments},
  author={Cao, Bryan Bo and Alali, Abrar and Liu, Hansi and Meegan, Nicholas and Gruteser, Marco and Dana, Kristin and Ashok, Ashwin and Jain, Shubham},
  booktitle={2022 19th Annual IEEE International Conference on Sensing, Communication, and Networking (SECON)},
  pages={19--27},
  year={2022},
  organization={IEEE}
}

@inproceedings{liu2023vifi,
  title={ViFi-Loc: Multi-modal pedestrian localization using GAN with camera-phone correspondences},
  author={Liu, Hansi and Lu, Hongsheng and Data, Kristin and Gruteser, Marco},
  booktitle={Proceedings of the 25th International Conference on Multimodal Interaction},
  pages={661--669},
  year={2023}
}

@inproceedings{papaioannou2015accurate,
  title={Accurate positioning via cross-modality training},
  author={Papaioannou, Savvas and Wen, Hongkai and Xiao, Zhuoling and Markham, Andrew and Trigoni, Niki},
  booktitle={Proceedings of the 13th ACM Conference on Embedded Networked Sensor Systems},
  pages={239--251},
  year={2015}
}

@inproceedings{alahi2015rgb,
  title={RGB-W: When vision meets wireless},
  author={Alahi, Alexandre and Haque, Albert and Fei-Fei, Li},
  booktitle={Proceedings of the IEEE International Conference on Computer Vision},
  pages={3289--3297},
  year={2015}
}

@inproceedings{strelow2001precise,
  title={Precise omnidirectional camera calibration},
  author={Strelow, Dennis and Mishler, Jeffrey and Koes, David and Singh, Sanjiv},
  booktitle={Proceedings of the 2001 IEEE Computer Society Conference on Computer Vision and Pattern Recognition. CVPR 2001},
  volume={1},
  pages={I--I},
  year={2001},
  organization={IEEE}
}

@article{liu2016sensor,
  title={Sensor selection for estimation with correlated measurement noise},
  author={Liu, Sijia and Chepuri, Sundeep Prabhakar and Fardad, Makan and Ma{\c{s}}azade, Engin and Leus, Geert and Varshney, Pramod K},
  journal={IEEE Transactions on Signal Processing},
  volume={64},
  number={13},
  pages={3509--3522},
  year={2016},
  publisher={IEEE}
}

@article{di2002counteraction,
  title={Counteraction of environmental disturbances of electronic nose data by independent component analysis},
  author={Di Natale, Corrado and Martinelli, Eugenio and D’Amico, Arnaldo},
  journal={Sensors and Actuators B: Chemical},
  volume={82},
  number={2-3},
  pages={158--165},
  year={2002},
  publisher={Elsevier}
}

@article{buhlmann2003boosting,
  title={Boosting with the L 2 loss: regression and classification},
  author={B{\"u}hlmann, Peter and Yu, Bin},
  journal={Journal of the American Statistical Association},
  volume={98},
  number={462},
  pages={324--339},
  year={2003},
  publisher={Taylor \& Francis}
}

@inproceedings{liu2020vision,
  title={Vision meets wireless positioning: Effective person re-identification with recurrent context propagation},
  author={Liu, Yiheng and Zhou, Wengang and Xi, Mao and Shen, Sanjing and Li, Houqiang},
  booktitle={Proceedings of the 28th ACM International Conference on Multimedia},
  pages={1103--1111},
  year={2020}
}

@inproceedings{xu2023uncovering,
  title={Uncovering the Missing Pattern: Unified Framework Towards Trajectory Imputation and Prediction},
  author={Xu, Yi and Bazarjani, Armin and Chi, Hyung-gun and Choi, Chiho and Fu, Yun},
  booktitle={Proceedings of the IEEE/CVF Conference on Computer Vision and Pattern Recognition},
  pages={9632--9643},
  year={2023}
}

@article{fujii2021two,
  title={A two-block rnn-based trajectory prediction from incomplete trajectory},
  author={Fujii, Ryo and Vongkulbhisal, Jayakorn and Hachiuma, Ryo and Saito, Hideo},
  journal={IEEE Access},
  volume={9},
  pages={56140--56151},
  year={2021},
  publisher={IEEE}
}

@inproceedings{zhang2023layout,
  title={Layout Sequence Prediction From Noisy Mobile Modality},
  author={Zhang, Haichao and Xu, Yi and Lu, Hongsheng and Shimizu, Takayuki and Fu, Yun},
  booktitle={Proceedings of the 31st ACM International Conference on Multimedia},
  pages={3965--3974},
  year={2023}
}

@article{zhao2017adaptive,
  title={Adaptive two-stage Kalman filter for SINS/odometer integrated navigation systems},
  author={Zhao, Hongsong and Miao, Lingjuan and Shao, Haijun},
  journal={The Journal of Navigation},
  volume={70},
  number={2},
  pages={242--261},
  year={2017},
  publisher={Cambridge University Press}
}

@inproceedings{liu2022vi,
  title={Vi-Fi: Associating Moving Subjects across Vision and Wireless Sensors},
  author={Liu, Hansi and Alali, Abrar and Ibrahim, Mohamed and Cao, Bryan Bo and Meegan, Nicholas and Li, Hongyu and Gruteser, Marco and Jain, Shubham and Dana, Kristin and Ashok, Ashwin and others},
  booktitle={2022 21st ACM/IEEE International Conference on Information Processing in Sensor Networks (IPSN)},
  pages={208--219},
  year={2022},
  organization={IEEE}
}

@article{martin2021jrdb,
  title={Jrdb: A dataset and benchmark of egocentric robot visual perception of humans in built environments},
  author={Martin-Martin, Roberto and Patel, Mihir and Rezatofighi, Hamid and Shenoi, Abhijeet and Gwak, JunYoung and Frankel, Eric and Sadeghian, Amir and Savarese, Silvio},
  journal={IEEE transactions on pattern analysis and machine intelligence},
  year={2021},
  publisher={IEEE}
}

@inproceedings{shi2018lstm,
  title={LSTM-based flight trajectory prediction},
  author={Shi, Zhiyuan and Xu, Min and Pan, Quan and Yan, Bing and Zhang, Haimin},
  booktitle={2018 International Joint Conference on Neural Networks (IJCNN)},
  pages={1--8},
  year={2018},
  organization={IEEE}
}

@inproceedings{tran2021goal,
  title={Goal-driven long-term trajectory prediction},
  author={Tran, Hung and Le, Vuong and Tran, Truyen},
  booktitle={Proceedings of the IEEE/CVF winter conference on applications of computer vision},
  pages={796--805},
  year={2021}
}

@inproceedings{giuliari2021transformer,
  title={Transformer networks for trajectory forecasting},
  author={Giuliari, Francesco and Hasan, Irtiza and Cristani, Marco and Galasso, Fabio},
  booktitle={2020 25th international conference on pattern recognition (ICPR)},
  pages={10335--10342},
  year={2021},
  organization={IEEE}
}

@inproceedings{rella2021decoder,
  title={Decoder fusion rnn: Context and interaction aware decoders for trajectory prediction},
  author={Rella, Edoardo Mello and Zaech, Jan-Nico and Liniger, Alexander and Van Gool, Luc},
  booktitle={2021 IEEE/RSJ International Conference on Intelligent Robots and Systems (IROS)},
  pages={5937--5943},
  year={2021},
  organization={IEEE}
}

@inproceedings{zhou2022hivt,
  title={Hivt: Hierarchical vector transformer for multi-agent motion prediction},
  author={Zhou, Zikang and Ye, Luyao and Wang, Jianping and Wu, Kui and Lu, Kejie},
  booktitle={Proceedings of the IEEE/CVF Conference on Computer Vision and Pattern Recognition},
  pages={8823--8833},
  year={2022}
}

@article{huang2006low,
  title={A low-order DGPS-based vehicle positioning system under urban environment},
  author={Huang, Jihua and Tan, H-S},
  journal={IEEE/ASME Transactions on mechatronics},
  volume={11},
  number={5},
  pages={567--575},
  year={2006},
  publisher={IEEE}
}

@article{hochreiter1997long,
  title={Long short-term memory},
  author={Hochreiter, Sepp and Schmidhuber, J{\"u}rgen},
  journal={Neural computation},
  volume={9},
  number={8},
  pages={1735--1780},
  year={1997},
  publisher={MIT press}
}

@article{chung2014empirical,
  title={Empirical evaluation of gated recurrent neural networks on sequence modeling},
  author={Chung, Junyoung and Gulcehre, Caglar and Cho, KyungHyun and Bengio, Yoshua},
  journal={arXiv preprint arXiv:1412.3555},
  year={2014}
}

@article{vaswani2017attention,
  title={Attention is all you need},
  author={Vaswani, Ashish and Shazeer, Noam and Parmar, Niki and Uszkoreit, Jakob and Jones, Llion and Gomez, Aidan N and Kaiser, {\L}ukasz and Polosukhin, Illia},
  journal={Advances in neural information processing systems},
  volume={30},
  year={2017}
}

@article{rumelhart1986learning,
  title={Learning Internal Representations by Error Propagation, Parallel Distributed Processing, Explorations in the Microstructure of Cognition, ed. DE Rumelhart and J. McClelland. Vol. 1. 1986},
  author={Rumelhart, David E and Hinton, Geoffrey E and Williams, Ronald J},
  journal={Biometrika},
  volume={71},
  pages={599--607},
  year={1986}
}

@article{kalman1960new,
  title={A New Approach to Linear Filtering and Prediction Problems},
  author={Kalman, Rudolf Emil},
  journal={Transactions of the ASME--Journal of Basic Engineering},
  volume={82},
  series={Series D},
  pages={35--45},
  year={1960}
}

@article{xia2025another,
  title={Another vertical view: A hierarchical network for heterogeneous trajectory prediction via spectrums},
  author={Xia, Beihao and Wong, Conghao and Xu, Duanquan and Peng, Qinmu and You, Xinge},
  journal={IEEE Transactions on Pattern Analysis and Machine Intelligence},
  year={2025},
  publisher={IEEE}
}

@article{zhang2023towards,
  title={Towards trajectory forecasting from detection},
  author={Zhang, Pu and Bai, Lei and Wang, Yuning and Fang, Jianwu and Xue, Jianru and Zheng, Nanning and Ouyang, Wanli},
  journal={IEEE Transactions on Pattern Analysis and Machine Intelligence},
  volume={45},
  number={10},
  pages={12550--12561},
  year={2023},
  publisher={IEEE}
}

@inproceedings{wong2025resonance,
  title={Resonance: Learning to Predict Social-Aware Pedestrian Trajectories as Co-Vibrations},
  author={Wong, Conghao and Zou, Ziqian and Xia, Beihao},
  booktitle={Proceedings of the IEEE/CVF International Conference on Computer Vision},
  pages={25788--25799},
  year={2025}
}

@inproceedings{wong2024socialcircle,
  title={Socialcircle: Learning the angle-based social interaction representation for pedestrian trajectory prediction},
  author={Wong, Conghao and Xia, Beihao and Zou, Ziqian and Wang, Yulong and You, Xinge},
  booktitle={Proceedings of the IEEE/CVF Conference on Computer Vision and Pattern Recognition},
  pages={19005--19015},
  year={2024}
}

@article{xia2022cscnet,
  title={CSCNet: Contextual semantic consistency network for trajectory prediction in crowded spaces},
  author={Xia, Beihao and Wong, Conghao and Peng, Qinmu and Yuan, Wei and You, Xinge},
  journal={Pattern Recognition},
  volume={126},
  pages={108552},
  year={2022},
  publisher={Elsevier}
}

@inproceedings{zhang2025vqtoken,
  title={Vqtoken: Neural discrete token representation learning for extreme token reduction in video large language models},
  author={Zhang, Haichao and Fu, Yun},
  booktitle={The Thirty-ninth Annual Conference on Neural Information Processing Systems},
  year={2025}
}

@article{zhang2025linkedout,
  title={LinkedOut: Linking World Knowledge Representation Out of Video LLM for Next-Generation Video Recommendation},
  author={Zhang, Haichao and Lu, Yao and Wang, Lichen and Li, Yunzhe and Chen, Daiwei and Xu, Yunpeng and Fu, Yun},
  journal={arXiv preprint arXiv:2512.16891},
  year={2025}
}

@article{zhang2025dense,
  title={Dense video understanding with gated residual tokenization},
  author={Zhang, Haichao and Chai, Wenhao and He, Shwai and Li, Ang and Fu, Yun},
  journal={arXiv preprint arXiv:2509.14199},
  year={2025}
}

@inproceedings{cao2022tagging,
  title={Tagging Vision with Smartphone Identities by Vision2Phone Translation},
  author={Cao, Bryan B and Alali, Abrar and Liu, Hansi and Meegan, Nicholas and Gruteser, Marco and Dana, Kristin and Ashok, Ashwin and Jain, Shubham},
  booktitle={IEEE International Conference on Sensing, Communication, and Networking September 2022},
  year={2022}
}

@inproceedings{cao2023vifit,
  title={Vifit: Reconstructing vision trajectories from imu and wi-fi fine time measurements},
  author={Cao, Bryan Bo and Alali, Abrar and Liu, Hansi and Meegan, Nicholas and Gruteser, Marco and Dana, Kristin and Ashok, Ashwin and Jain, Shubham},
  booktitle={Proceedings of the 3rd ACM MobiCom Workshop on Integrated Sensing and Communications Systems},
  pages={13--18},
  year={2023}
}

@inproceedings{wong2022view,
  title={View vertically: A hierarchical network for trajectory prediction via fourier spectrums},
  author={Wong, Conghao and Xia, Beihao and Hong, Ziming and Peng, Qinmu and Yuan, Wei and Cao, Qiong and Yang, Yibo and You, Xinge},
  booktitle={European Conference on Computer Vision},
  pages={682--700},
  year={2022},
  organization={Springer}
}

@article{wong2023msn,
  title={MSN: Multi-style network for trajectory prediction},
  author={Wong, Conghao and Xia, Beihao and Peng, Qinmu and Yuan, Wei and You, Xinge},
  journal={IEEE Transactions on Intelligent Transportation Systems},
  volume={24},
  number={9},
  pages={9751--9766},
  year={2023},
  publisher={IEEE}
}

@article{xie2024pedestrian,
  title={Pedestrian trajectory prediction based on social interactions learning with random weights},
  author={Xie, Jiajia and Zhang, Sheng and Xia, Beihao and Xiao, Zhu and Jiang, Hongbo and Zhou, Siwang and Qin, Zheng and Chen, Hongyang},
  journal={IEEE Transactions on Multimedia},
  volume={26},
  pages={7503--7515},
  year={2024},
  publisher={IEEE}
}

@inproceedings{Jianglin2025,
  title={The Indra Representation Hypothesis for Multimodal Alignment},
  author={Jianglin Lu and Hailing Wang and Kuo Yang and Yitian Zhang and Simon Jenni and Yun Fu},
  booktitle={The Thirty-ninth Annual Conference on Neural Information Processing Systems},
  year={2025},
}

@inproceedings{lu2025representation,
  title={Representation Potentials of Foundation Models for Multimodal Alignment: A Survey},
  author={Jianglin Lu and Hailing Wang and Yi Xu and Yizhou Wang and Kuo Yang and Yun Fu},
  booktitle={Proceedings of the 2025 Conference on Empirical Methods in Natural Language Processing},
  year={2025},
}

@inproceedings{zhangstable,
  title={Stable Part Diffusion 4D: Multi-View RGB and Kinematic Parts Video Generation},
  author={Zhang, Hao and Yao, Chun-Han and Donn{\'e}, Simon and Ahuja, Narendra and Jampani, Varun},
  booktitle={The Thirty-ninth Annual Conference on Neural Information Processing Systems},
year={2025}
}

@inproceedings{zhang2024open,
  title={Open-nerf: Towards open vocabulary nerf decomposition},
  author={Zhang, Hao and Li, Fang and Ahuja, Narendra},
  booktitle={Proceedings of the IEEE/CVF Winter Conference on Applications of Computer Vision},
  pages={3456--3465},
  year={2024}
}

@inproceedings{feng2025videoorion,
  title={VideoOrion: Tokenizing Object Dynamics in Videos},
  author={Feng, Yicheng and Li, Yijiang and Zhang, Wanpeng and Zheng, Sipeng and Luo, Hao and Yue, Zihao and Lu, Zongqing},
  booktitle={Proceedings of the IEEE/CVF International Conference on Computer Vision},
  pages={20401--20412},
  year={2025}
}

@inproceedings{zhangpixels,
  title={From Pixels to Tokens: Byte-Pair Encoding on Quantized Visual Modalities},
  author={Zhang, Wanpeng and Xie, Zilong and Feng, Yicheng and Li, Yijiang and Xing, Xingrun and Zheng, Sipeng and Lu, Zongqing},
  booktitle={The Thirteenth International Conference on Learning Representations},
  year={2025}
}

@inproceedings{li2025core,
  title={Core Knowledge Deficits in Multi-Modal Language Models},
  author={Li, Yijiang and Gao, Qingying and Zhao, Tianwei and Wang, Bingyang and Sun, Haoran and Lyu, Haiyun and Hawkins, Robert D and Vasconcelos, Nuno and Golan, Tal and Luo, Dezhi and others},
  booktitle={International Conference on Machine Learning},
  pages={34379--34409},
  year={2025},
  organization={PMLR}
}

@inproceedings{zhang2025unified,
  title={Unified multimodal understanding via byte-pair visual encoding},
  author={Zhang, Wanpeng and Feng, Yicheng and Luo, Hao and Li, Yijiang and Yue, Zihao and Zheng, Sipeng and Lu, Zongqing},
  booktitle={Proceedings of the IEEE/CVF International Conference on Computer Vision},
  pages={12976--12986},
  year={2025}
}

@article{zhang2026thinkjepa,
  title={ThinkJEPA: Empowering Latent World Models with Large Vision-Language Reasoning Model},
  author={Zhang, Haichao and Li, Yijiang and He, Shwai and Nagarajan, Tushar and Chen, Mingfei and Lu, Jianglin and Li, Ang and Fu, Yun},
  journal={arXiv preprint arXiv:2603.22281},
  year={2026}
}

@inproceedings{nikhil2018convolutional,
  title={Convolutional neural network for trajectory prediction},
  author={Nikhil, Nishant and Tran Morris, Brendan},
  booktitle={Proceedings of the European Conference on Computer Vision (ECCV) Workshops},
  pages={0--0},
  year={2018}
}

@article{he2025understanding,
  title={Understanding and Harnessing Sparsity in Unified Multimodal Models},
  author={He, Shwai and Deng, Chaorui and Li, Ang and Yan, Shen},
  journal={arXiv preprint arXiv:2512.02351},
  year={2025}
}

@inproceedings{girgis2022latent,
  title={Latent Variable Sequential Set Transformers for Joint Multi-Agent Motion Prediction},
  author={Girgis, Roger and Golemo, Florian and Codevilla, Felipe and Weiss, Martin and D'Souza, Jim Aldon and Kahou, Samira Ebrahimi and Heide, Felix and Pal, Christopher},
  booktitle={International Conference on Learning Representations},
  year={2022},
  url={https://openreview.net/forum?id=Dup\_dDqkZC5}
}

@inproceedings{liu2021lost,
  title={Lost and Found! associating target persons in camera surveillance footage with smartphone identifiers},
  author={Liu, Hansi and Alali, Abrar and Ibrahim, Mohamed and Li, Hongyu and Gruteser, Marco and Jain, Shubham and Dana, Kristin and Ashok, Ashwin and Cheng, Bin and Lu, Hongsheng},
  booktitle={Proceedings of the 19th Annual International Conference on Mobile Systems, Applications, and Services},
  pages={499--500},
  year={2021}
}
}



\begin{IEEEbiography}[{\includegraphics[width=1in,height=1.25in,clip,keepaspectratio]{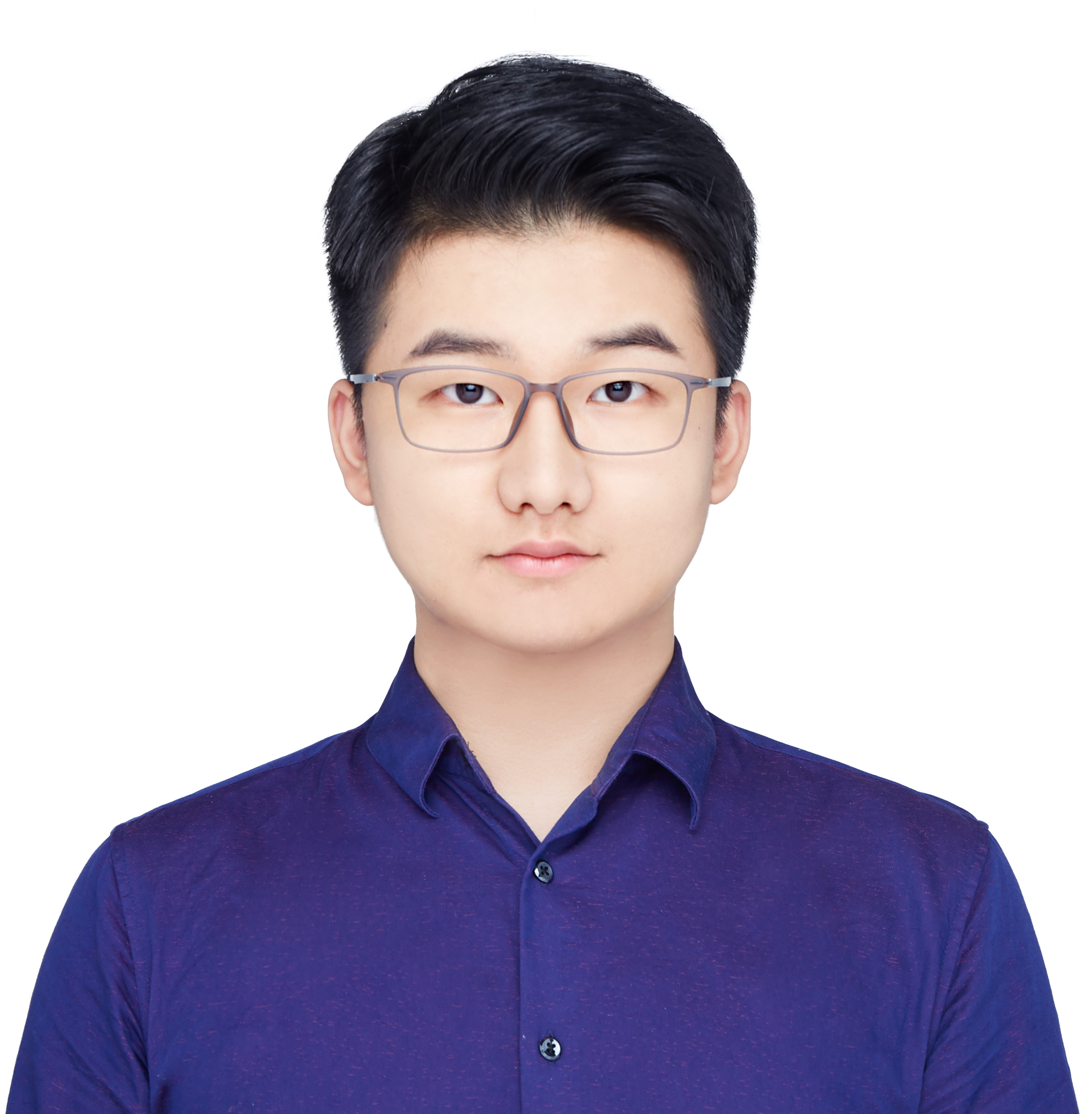}}]{Haichao Zhang, Student Member, IEEE}
is a Ph.D. candidate in Computer Engineering with the Department of Electrical and Computer Engineering at Northeastern University, USA. He received the M.S. degree from Zhejiang University, China, in 2021. His research interests lie in generative AI for computer vision, with a particular focus on vision--language models, video understanding and generation, world models, and trajectory and motion prediction and planning. His work centers on token-efficient dense video understanding and world modeling for applications in AI-assisted content creation, autonomous systems, and robotics. He has held research internships at Meta Reality Labs Research, LinkedIn Video AI, Amazon AWS AI Labs, and Tencent. He serves as an area chair for the ICLR ES-Reasoning Workshop and the CVPR MMRAgI Workshop, and as a reviewer for NeurIPS, IJCAI, ACM MM, ICLR, ECCV, ICCV, ICML, AISTATS, TPAMI, and TIP.
\end{IEEEbiography}

\begin{IEEEbiography}[{\includegraphics[width=1in,height=1.25in,clip,keepaspectratio]{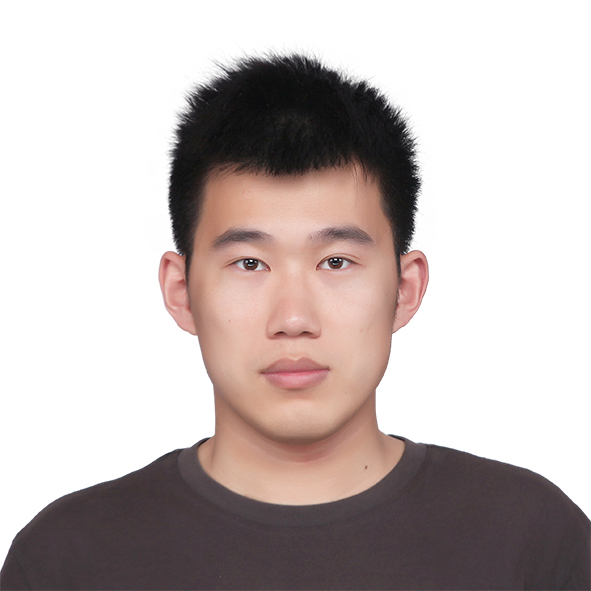}}]{Yi Xu, Student Member, IEEE} received the BS and MS degrees from Xi’an Jiaotong University in 2017 and 2020 respectively. Currently he is working toward the PhD degree with Northeastern University. His current research interests are machine learning, computer vision, pattern recognition, and their applications to intelligent systems.
\end{IEEEbiography}

\begin{IEEEbiography}[{\includegraphics[width=1in,height=1.25in,clip,keepaspectratio]{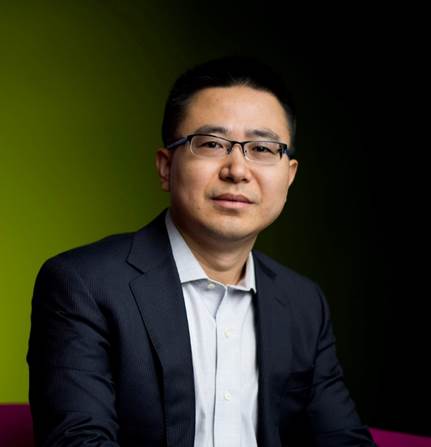}}]{Yun (Raymond) Fu, IEEE Fellow}
is a Member of Academia Europaea and the European Academy of Sciences and Arts, a Fellow of the National Academy of Inventors (NAI), IEEE, ACM, AAAI, AAAS, AIMBE, IAPR, OSA, SPIE, and AAIA. He received the Ph.D. degree in Electrical and Computer Engineering from the University of Illinois at Urbana–Champaign. He is a Distinguished Professor affiliated with both the College of Engineering and the Khoury College of Computer Sciences at Northeastern University. He has authored more than 500 scientific publications and holds over 40 patents. His honors include seven Prestigious Young Investigator Awards (from NAE, ONR, ARO, IEEE, INNS, UIUC, and the Grainger Foundation), twelve Best Paper Awards (from IEEE, ACM, IAPR, SPIE, and SIAM), and major industrial awards from Google, Amazon, Samsung, Adobe, JPMorgan Chase, NEC, Snap, Cisco, Toyota, MERL, PicsArt, Konica Minolta, Zebra, and MathWorks. He is a Lifetime Senior Member of AAAI and the Institute of Mathematical Statistics, and has served on the ACM Future of Computing Academy, the Global Young Academy, AAAS, and INNS. 
\end{IEEEbiography}

\end{document}